\documentclass{article}

\usepackage{microtype}
\usepackage{graphicx}
\usepackage{subfigure}
\usepackage{booktabs} 

\PassOptionsToPackage{hyphens}{url}
\usepackage{hyperref}

\usepackage[accepted]{icml2024}

\usepackage{amsmath}
\usepackage{amssymb}
\usepackage{mathtools}
\usepackage{amsthm}

\usepackage[capitalize,noabbrev]{cleveref}
\usepackage{xcolor}

\newsavebox{\coloredquotationbox}
\newenvironment{coloredquotation}
 {%
  \begin{trivlist}
  \begin{lrbox}{\coloredquotationbox}
  \begin{minipage}{\dimexpr\linewidth-2\fboxsep}
 }
 {%
  \end{minipage}
  \end{lrbox}
  \item\relax
  \parbox{\linewidth}{
    \begingroup
    \color[RGB]{224,215,188}%
    \hrule
    \color[RGB]{249,245,233}%
    \hrule
    \color[RGB]{224,215,188}%
    \hrule
    \endgroup
    \colorbox[RGB]{249,245,233}{\usebox{\coloredquotationbox}}\par\nointerlineskip
    \begingroup
    \color[RGB]{224,215,188}%
    \hrule
    \color[RGB]{249,245,233}%
    \hrule
    \color[RGB]{224,215,188}%
    \hrule
    \endgroup
  }
  \end{trivlist}
 }

\theoremstyle{plain}

\theoremstyle{definition}

\theoremstyle{remark}

\usepackage[textsize=tiny]{todonotes}

\usepackage{enumitem}
\usepackage{tikz}
\usepackage{pgfplots}
\usepackage{xspace}
\usepackage[frozencache,cachedir=.]{minted}

\usepackage{xcolor} 
\definecolor{LightGray}{gray}{0.7}

\usepackage[T1]{fontenc}

\usepackage[utf8]{inputenc}

\usepackage{microtype}
\usepackage{makecell}

\usepackage{array}
\usepackage{booktabs}
\usepackage{soul}
\usepackage{url}
\usepackage[utf8]{inputenc}
\usepackage{graphicx}
\usepackage{amsmath}
\usepackage{booktabs}
\usepackage{colortbl}
\usepackage{multirow}
\usepackage{float}	
\usepackage{dblfloatfix}
\usepackage{setspace}  

\usepackage{xcolor}	

\usepackage{cleveref}

\definecolor{ForestGreen}{rgb}{0.13, 0.55, 0.13}
\definecolor{skyblue}{HTML}{46C5DD}	

\newcommand{\eat}[1]{}

\usepackage{quoting}
\newenvironment{ite}{                     
     \parskip 0cm \begin{itemize} \parskip 0cm \parsep 0cm \itemsep 0cm \topsep 0cm}{
        \end{itemize}} 

\usepackage[]{mdframed}
\usepackage{framed}

\usepackage{ctable} 
\usepackage{multicol}

\renewcommand{\cite}{\citep}

\definecolor{lightgray}{gray}{0.9}
\definecolor{Box1Color}{RGB}{227, 236, 246}
\definecolor{Box2Color}{RGB}{248, 220, 225}
\definecolor{Box3Color}{RGB}{255, 238, 224}
\definecolor{cbBlue}{RGB}{0, 114, 178}
\definecolor{cbOrange}{RGB}{240, 228, 66}
\definecolor{cbGreen}{RGB}{0, 158, 115}
\definecolor{cbRed}{RGB}{213, 94, 0}
\definecolor{cbPurple}{RGB}{204, 121, 167}
\definecolor{cbSkyBlue}{RGB}{86, 180, 233}
\definecolor{cbGray}{RGB}{128, 128, 128}
\definecolor{CBF1}{RGB}{255,99,132}  
\definecolor{CBF2}{RGB}{54,162,235}  
\definecolor{CBF3}{RGB}{255,206,86}  
\definecolor{CBF4}{RGB}{75,192,192}  
\definecolor{CBF5}{RGB}{153,102,255} 
\definecolor{CBF1b}{RGB}{205,89,112}  
\definecolor{CBF2b}{RGB}{44,142,215}  
\definecolor{CBF5b}{RGB}{133,92,225}  
\definecolor{PromptBgColor}{RGB}{255, 238, 224}

\usepackage{amssymb}
\usepackage{pifont}

\newcommand{\mypara}[1]{\textbf{#1}\hspace{0.7em}}

\newcommand{\osod}{\textsc{DataVoyager}}

\icmltitlerunning{Position Paper: Data-driven Discovery with Large Generative Models}

\begin{document}

\twocolumn[
\icmltitle{Data-driven Discovery with Large Generative Models}



\icmlsetsymbol{equal}{*}

\begin{icmlauthorlist}
\icmlauthor{Bodhisattwa Prasad Majumder}{equal,ai2}
\icmlauthor{Harshit Surana}{equal,openlocus}
\icmlauthor{Dhruv Agarwal}{umass}
\icmlauthor{Sanchaita Hazra}{utah}\\
\icmlauthor{Ashish Sabharwal}{ai2}
\icmlauthor{Peter Clark}{ai2}
\end{icmlauthorlist}


\icmlaffiliation{ai2}{Allen Institute for AI}
\icmlaffiliation{openlocus}{OpenLocus}
\icmlaffiliation{umass}{University of Massachusetts Amherst}
\icmlaffiliation{utah}{University of Utah}

\icmlcorrespondingauthor{Bodhisattwa Prasad Majumder}{bodhisattwam@allenai.org}
\icmlcorrespondingauthor{Harshit Surana}{harshit@openlocus.ai}


\icmlkeywords{Machine Learning, ICML}

\vskip 0.2in
]



\printAffiliationsAndNotice{\icmlEqualContribution} 

\newcount\Comments  
\Comments=1  
\definecolor{darkgreen}{rgb}{0,0.5,0}
\definecolor{darkred}{rgb}{0.7,0,0}
\definecolor{teal}{rgb}{0.1,0.6,0.7}
\definecolor{blue}{rgb}{0.0,0.1,0.9}
\definecolor{orange}{rgb}{1.,0.7,0.0}
\definecolor{lightblue}{rgb}{0.70, 0.80, 0.89}
\definecolor{violet}{rgb}{0.50, 0.16, 0.88}
\newcommand{\kibitz}[2]{\ifnum\Comments=1{{\textcolor{#1}{\textsf{\footnotesize [#2]}}}}\fi}
\newcommand{\dhruv}[1]{\kibitz{darkred}{Dhruv: #1}}
\newcommand{\bodhi}[1]{\kibitz{blue}{Bodhi: #1}}
\newcommand{\harshit}[1]{\kibitz{teal}{Harshit: #1}}

\newcommand{\ashish}[1]{\kibitz{darkgreen}{Ashish: #1}}
\newcommand{\pete}[1]{\kibitz{orange}{PeteC: #1}}
\newcommand{\yoav}[1]{\kibitz{violet}{Yoav: #1}}

\begin{abstract}
With the accumulation of data at an unprecedented rate, its potential to fuel scientific discovery is growing exponentially. This position paper urges the Machine Learning (ML) community to exploit the capabilities of large generative models (LGMs) to develop automated systems for end-to-end \emph{data-driven discovery}---a paradigm encompassing the search and verification of hypotheses purely from a set of provided datasets, without the need for additional data collection or physical experiments. We first outline several desiderata for an ideal data-driven discovery system. Then, through \osod, a proof-of-concept utilizing GPT-4, we demonstrate how LGMs fulfill several of these desiderata---a feat previously unattainable---while also highlighting important limitations in the current system that open up opportunities for novel ML research. We contend that achieving accurate, 
reliable, and robust end-to-end discovery systems solely through the current capabilities of LGMs is challenging. We instead advocate for fail-proof tool integration, along with active user moderation through feedback mechanisms, to foster data-driven scientific discoveries with efficiency and reproducibility.
\end{abstract}

\section{Introduction}


The deluge of data collected in the digital age by advanced scientific instruments, sensors, and computational techniques has marked a transformative change in the process and pace of scientific discovery \cite{anderson2008end, ramakrishnan1999data, jumper2021highly}. 
This acceleration, however, paints a paradoxical scenario---while rapid development indicates the advancement of knowledge, it simultaneously poses significant challenges for scientists to absorb new findings, navigate interconnections, formulate novel hypotheses, and arrive at meaningful conclusions \cite{bianchini2022artificial}. To facilitate future scientific progress, it is, therefore, imperative to develop automated systems that are capable of continuous ingestion, creative generation, and analytical reasoning at a massive scale.

\begin{figure*}[t] 
\centering
\includegraphics[trim= 20 28 20 20,clip, width=\linewidth]{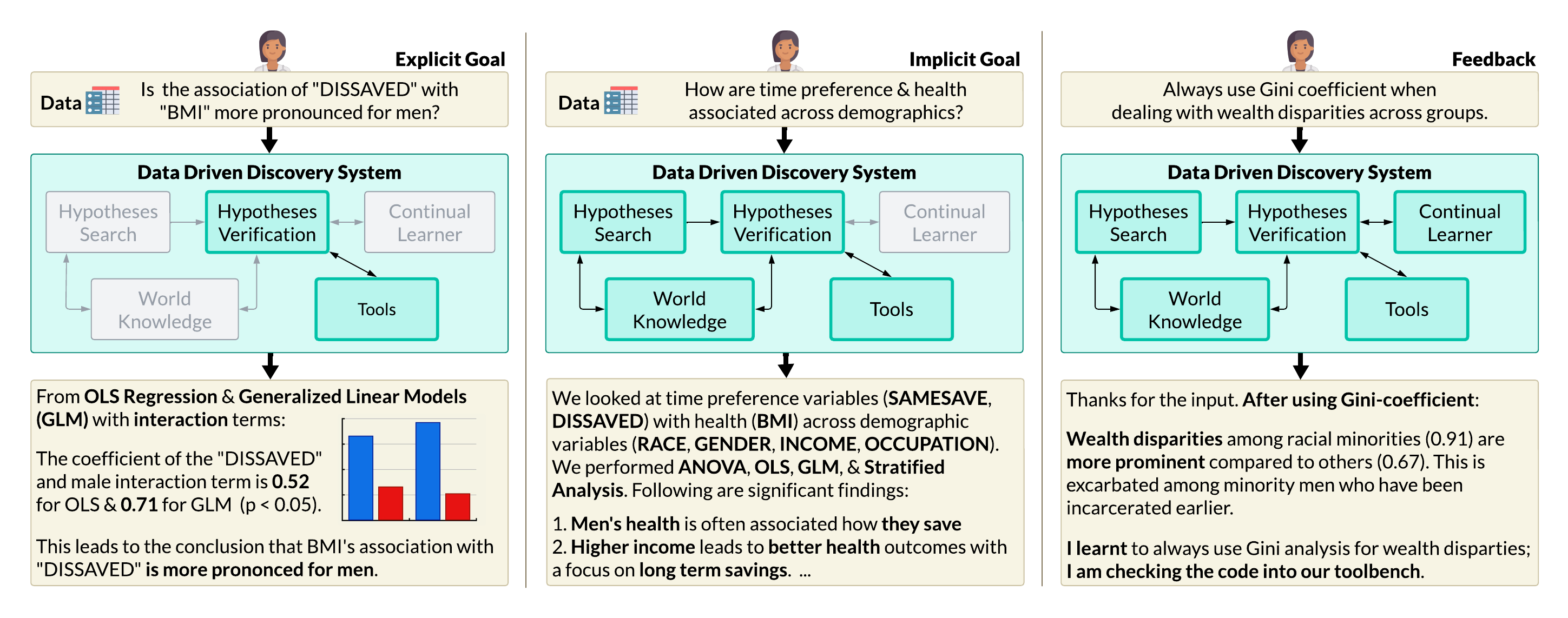}
\vspace{-1em}
 \caption{{\textbf{A blueprint flow} demonstrating ideal workflows for data-driven discovery. \textbf{Left:} User asks an explicit question around a particular line of inquiry or hypothesis. \textbf{Middle:} The user can also ask a broad and partially-defined
 high-level question, where the system must figure out the appropriate datasets, data transformations, variables, a list of possible hypotheses, and their verification. In this example, the system maps time preference and health outcomes to exact variables, runs the analysis across appropriate demographic cuts, and then shares the significant findings for further exploration and verification. \textbf{Right:} The user can provide follow-up feedback at any time and the continual learner will learn from it while providing updated experiments and results.}}
    \label{fig:blueprint}
\end{figure*}

Developing an end-to-end discovery system is challenging. 
Previous works have either severely lacked the requisite computational power
\citep{Langley1981DataDrivenDO, Langley1984TheSF, Langley1983RediscoveringCW}, developed domain-specific bespoke methodologies (e.g., AlphaFold; \citet{jumper2021highly}), or involved substantial human intervention (e.g., wet lab experiments) thus not qualifying as autonomous end-to-end (CoScientist; \citet{Boiko2023AutonomousCR}). 
In this position paper, we argue that a focus 
on \textbf{data-driven discovery} using \textbf{large generative models (LGMs)} 
addresses each of these prior shortcomings and presents a practical first step towards the goal of an end-to-end system for automating the scientific process. 
Following \citet{10.1145/360018.360022}, we define this paradigm
as a heuristic search framework that aims to describe a given set of observations by uncovering the laws that govern its data-generating process.

For example, consider the flow described in Figure~\ref{fig:blueprint}. Given a dataset of socio-economic variables collected from a set of respondents, a user might formulate a hypothesis about the relationship between the BMI of a subset of the respondents and their financial behavior (variables present in the dataset; \emph{top-left}). 
A data-driven discovery system should be able to automatically generate a verification plan and execute multiple steps of statistical tests (e.g., OLS, GLM) over the provided data to confirm or reject the hypothesis (\emph{bottom-left}). 
Alternatively, a user might only provide a high-level research question, such as specifying the domains of interest (i.e., finance and health; \emph{top-middle}). 
In this scenario, a discovery system must first identify the relevant variables and then search the space of plausible hypotheses to generate and verify interesting questions conditioned on the provided data and existing world knowledge (\emph{bottom-middle}). Finally, users may have diverse information-seeking needs necessitating the ability to provide feedback to the system, such as in using a particular statistical methodology for certain types of data during automatic verification (\emph{top-right}). An automated discovery system must accommodate and persist such feedback in order to recover from mistakes and accurately handle future queries (\emph{bottom-right}).

While our ultimate goal encompasses the full spectrum of scientific inquiry, we focus first on end-to-end discovery from observational or experimental data
for two reasons:
(1) an abundance of large-scale 
datasets that would benefit highly from
automated discovery; and (2) the practicality of automated verification 
enabled by 
data without the need for additional data collection\footnote{In contrast to hypothesis verification in the physical sciences, which often require wet lab experiments and where erroneous automation may lead to false discoveries \citep{leeman2024challenges}.}.

We identify two main challenges to automating data-driven discovery---(1) \emph{hypothesis search}: the effective consumption of provided data and existing knowledge to devise novel hypotheses, and (2) \emph{hypothesis verification}: the evaluation of the generated hypotheses for rapid iteration and continual discovery. 
A successful solution must further be able to generate and follow complex plans, execute diverse analytical tests, and parse through the abundant heterogeneity in real-world data. 
With the unprecedented success of LGMs operating on multiple modalities such as language \cite{Achiam2023GPT4TR, Touvron2023Llama2O}, code \cite{Liu2023IsYC, Li2022CompetitionlevelCG}, and images \cite{Achiam2023GPT4TR, Liu2023VisualIT},
we argue that it is now practical to build such a solution that can effectively tackle both challenges. 

\mypara{Hypothesis Search.} The scientific process typically begins with the construction of a proposed hypothesis based on prior knowledge and exploratory observations regarding some phenomenon of interest. 
For example, discovering new insights from publicly available National Longitudinal Surveys\footnote{\url{https://www.bls.gov/nls/}} will require prioritizing unexplored hypotheses over already verified results.
However, it is non-trivial what an optimal method for such a prioritized search might be. 

\begin{figure*}[t!]
    \centering
    \includegraphics[trim= 30 25 30 50,clip,width=\linewidth]{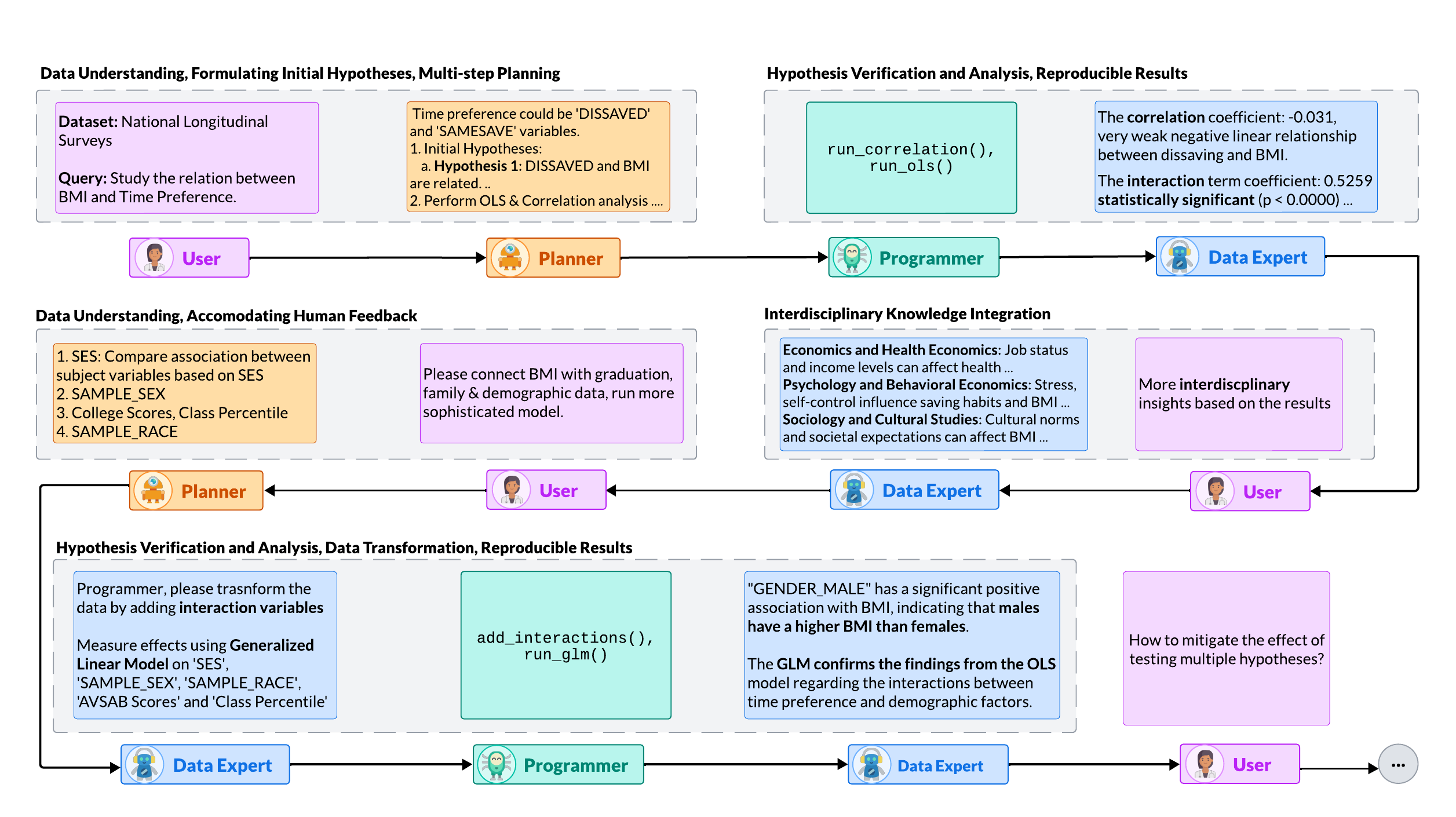}
    \vspace{-1em}
    \caption{\textbf{An example workflow} of \osod{}. Starting from a user-provided dataset and a high-level query, it navigates through cycles of hypothesis generation, validation, and analysis to uncover complex insights. See \textbf{all examples in Appendix} for full understanding. 
    }
    \label{fig:osod_fig1}
\end{figure*}

Foremost, we may ask whether the search should be driven by an extrinsic goal---a user-defined objective, a high-level research question, or a set of variables of interest. This setting might involve using algorithms that guide the search process using objective-gradients \cite{Weitzman1978OptimalSF} that identify variables and models that directly, or \emph{greedily}, optimize the extrinsic goal.
We argue that LGMs, with their massive, web-scale pre-training, possess both the necessary priors and the ability to handle heterogeneity, to help guide such a goal-driven search for relevant hypotheses.

It can also be argued that goal-driven approaches may not yield desired outcomes, particularly when dealing with open-ended questions, where the search is often susceptible to capture by local optima \citep{whitley1991fundamental, bengio2009curriculum}.
Drivers for search might then be intrinsic metrics 
\citep{oudeyer2007intrinsic}, such as diversity \citep{eysenbach2018diversity, Agarwal2023BringYO, trinh2024solving}, interestingness (or curiosity) \citep{pathak2017curiosity, zhang2023omni}, or information gain \citep{hennig2012entropy, houthooft2016vime}, that do not optimize for 
a user-defined extrinsic goal
but instead
encourage open-ended creativity and, eventually, serendipitous discovery \citep{foster2003serendipity, 10.5555/1324807, stanley2017open}. 
Here, too, LGMs present a solution, for instance, in estimating the novelty or likelihood of hypotheses in the search space.


\vfill

\mypara{Hypothesis Verification.} With a set of plausible hypotheses identified, it is next required to subject each claim through detailed inspection, often via a series of empirical evaluations and statistical tests, to determine veracity, which is highly tractable and could be fail-proof in data-driven discovery. This might involve selecting which analyses or statistical tests to run, transforming raw data into a format admissible for each test, handling missing or erroneous data, generating code to execute the tests, and finally analyzing the test results.
Given the surge of recent advancements in language modeling capabilities, including instruction-following \citep{wei2022finetuned}, tool use \citep{schick2023toolformer}, program synthesis \citep{Wang2023VoyagerAO, Agarwal2023BringYO}, planning \citep{Majumder2023CLINAC}, and orchestration \cite{Hou2023LargeLM},
we argue that LGM agents present a promising solution for automating hypothesis evaluation. 

The availability of these capabilities, however, must not be seen as a panacea. (1) LGMs often hallucinate, leading to incorrect insights that may not be grounded in the data. (2) LGMs have limited or no ``System-2'' reasoning 
\citep{kahneman2011thinking, lecun2022path, kambhampati2024llms}, thus necessitating additional scaffolding in order to utilize them for long-horizon tasks. (3) LGMs demonstrate subpar performance in the long tail, thus making their successful application in interfacing with external and domain-specific tools a major challenge to overcome. (4) Finally, LGMs are notoriously challenging to align and steer based on human feedback \cite{Wolf2023FundamentalLO}, a crucial component for reliable and useful scientific discovery.

\begin{figure*}[t!]
    \centering
    \includegraphics[trim= 35 500 35 60,clip, width=0.95\linewidth]{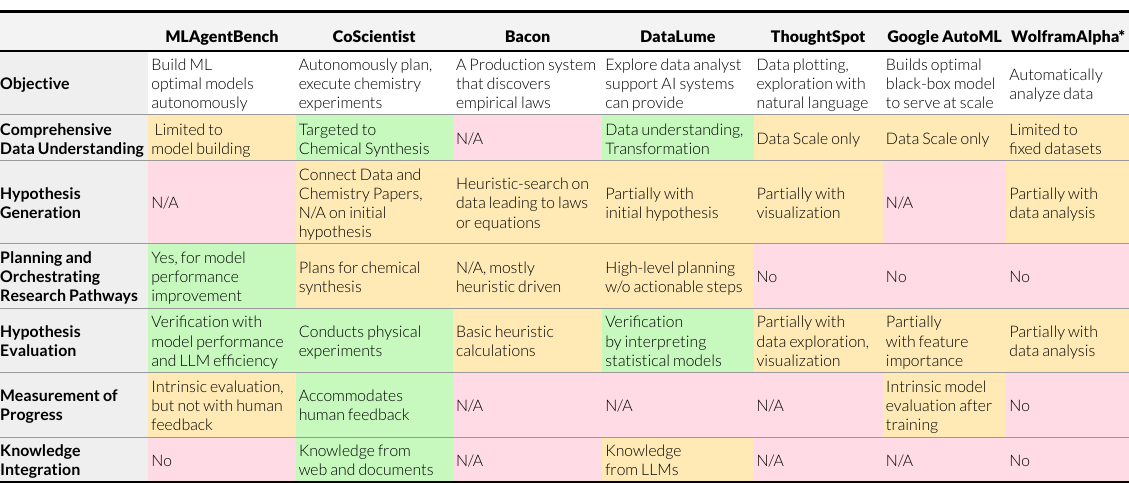}
    \vspace{-0.8em}
    \caption{\textbf{Survey} across several dimensions of a proposed data discovery system for several existing automated and semi-automated data analysis and discovery systems such as: MLAgentBench \cite{Huang2023BenchmarkingLL}, CoScientist \cite{Boiko2023AutonomousCR}, Bacon \cite{Langley1977BACONAP}, DataLume \cite{Gu2023HowDD}, ThoughtSpot (\url{thoughtspot.com}), Google AutoML (\url{cloud.google.com/automl}), and Automatic Analysis* from WolframAlpha (\url{wolframalpha.com/examples/pro-features/data-input}). 
    }
    \label{fig:osod_comparison}
\end{figure*}

We envision a blueprint of a data-driven discovery system in \Cref{fig:blueprint} that allows researchers to ingest datasets, search and verify hypotheses using fail-proof tools, and consult literature to surface novel insights. Our survey in \Cref{fig:osod_comparison} indicates the lack of systems capable of automated and robust data-driven discovery, with existing systems partially covering desired functionalities. To tackle this, we argue:
\begin{enumerate}[topsep=0pt,itemsep=0pt,parsep=0pt,partopsep=0pt]
    \item Automated data-driven discovery warrants research attention owing to the abundance of (public or private) data and its tractable challenges (hypothesis search and verification) as opposed to discoveries requiring laborious data collection or physical experiments.
    \item LGMs present an incredible potential to realize several properties of an ideal data-driven discovery system, such as knowledge-driven hypothesis search or tool usage to verify hypotheses---creating new avenues for ongoing efforts in the ML community on code generation, planning, and program synthesis.
    \item LGMs are \emph{not} all we need. Interfacing with fail-proof tools and inference-time functions, catering to domains and long tail with user moderation, is required to have an accurate, 
    reliable, and robust data-driven discovery system capable of advancing scientific progress with speed and reproducibility.
\end{enumerate}

\section{\osod{}: A Proof of Concept}

As a proof of concept, we borrow a well-studied role-based multi-agent architecture \cite{Liu2023AgentBenchEL, Zhou2023SOTOPIAIE} 
powered by GPT-4 \cite{Achiam2023GPT4TR}, a state-of-the-art language model, 
to build \osod{}---a 
system that can semantically understand a dataset, programmatically explore verifiable hypotheses using the available data, run basic statistical tests (e.g., correlation and regression analyses) by invoking pre-defined functions or generating code snippets, and finally analyze the output with detailed analyses. \osod is meant to represent a \emph{baseline system} that utilizes existing functionalities of 
GPT-4,
such as function calling, code generation, and language generation.

We envision any data-driven discovery system to be capable of operating in either of the following two settings. (1) \emph{Fully-autonomous}: using only the dataset and its metadata as the input. In this case, the system should consider the full hypothesis space
for search and verification. (2) \emph{User-guided}: combining the dataset with a (natural language) query stating a high-level objective to narrow down the hypothesis search space, akin to goal-directed agents \cite{Majumder2023CLINAC}. \osod{} can operate in both settings.
The core components of our system consist of specialized agents that are designed to manage different aspects of the data-driven discovery process as well as structured functions or programs that help analyze the data in specific ways via function calling.
We employ the AutoGen framework\footnote{\url{https://microsoft.github.io/autogen/}} that allows agents to communicate in arbitrary order dependent on the context.
Following is a brief description of all agents used in \osod{} (more in \Cref{fig:agent-structure}):
\vfill
\begin{ite}
    \item \textbf{Planner:} Interprets the user query and generates a comprehensive, structured plan to achieve it or, in the autonomous setting, generates an additional dataset exploration plan. The plan is then decomposed into executable sub-tasks and delegated to the relevant agents. 
    \item \textbf{Programmer:} Performs data transformations, filtering, and specialized coding for domain-specific analyses according to the generated plan. It can also call structured, pre-defined functions with relevant arguments to make execution fail-proof.\footnote{We develop several functions (e.g., statistical analysis tools based on datatype, python shell execution tools) for robustness.}
    \item \textbf{Data Expert:} Interprets the results generated by the programmer, extracting insights, connecting interdisciplinary knowledge, and formulating conclusions.
    \item \textbf{Critic:} Evaluates the analyses and provides constructive feedback on analytical methods and execution.
    \item \textbf{User Proxy:} Facilitates on-demand human feedback. A user can steer the discovery process towards an objective, rectify errors, and prevent off-course explorations.
\end{ite}
\vfill




\section{Towards Data-driven Discovery Systems}




In this section, we first outline a set of desired \textbf{functionalities} for a data-driven discovery system. Using these functionalities, and armed with our baseline system \osod{} along with evidence from the literature, 
we demonstrate extensive support towards our positions 2 and 3. 
Functionalities such as data understanding, hypothesis generation, multi-step planning, and interdisciplinary knowledge integration provide evidence that a 
system (\osod{}) powered by a state-of-the-art LGM shows promise for ideal data-driven discovery, an ability not previously achievable before the wide adoption of LGMs. 
On the other hand, 
functionalities such as data transformation, scale, hypothesis verification, accommodating human feedback, and $p$-hacking proof confirm that LGMs alone are insufficient. Integrating robust tools to execute at scale and user-centric interventions is crucial for a tractable data-driven discovery system.


\subsection{Comprehensive Data Understanding}
\mypara{Data Understanding.} 
Understanding data forms the core of data-driven discovery and involves processing variables semantically as well as programmatically \cite{Ristoski2016SemanticWI}. The former involves understanding how 
the data was collected or acquired, grounded in the domain knowledge,
as well as the semantic relationship between the variables present in the data. The latter involves understanding the datatype of each variable and the values they can take. Progress in database query generation \cite{Sun2023SQLPaLMIL}, commonsense reasoning on symbolic spaces \cite{Qiu2023PhenomenalYP}, and unsupervised KGQA \citep{Agarwal2023BringYO} indicate reason for optimism for the use of LGMs for data understanding.

For example, \citet{smith2005time} explored the link between time preference and BMI from the National Longitudinal Surveys using several variables indicating the saving behavior of the respondents. To replicate this from scratch, a discovery system must understand the difference between the variable measuring if respondents withdrew more money from savings than they put in (DISSAVED) and the variable indicating if they have no savings or unchanged savings from the previous year (SAMESAVE)\footnote{Time preference reflects how individuals value present over future benefits. A lower time preference can lead to higher savings, better food consumption, and thus a healthier BMI in the future.}. 
Here, \osod{}'s LGM-based planner correctly identifies variables related to time preference: 

\begin{coloredquotation}
To examine the effects of time preference on individuals, we need to understand the variables in the dataset that relate to time preference. In the provided dataset, the variables \textbf{\texttt{DISSAVED}} and \textbf{\texttt{SAMESAVE}} seem to be related to time preference (...) \textit{Full example:} \Cref{fig:data_understanding_1}
\end{coloredquotation}

While it works for this example, a comprehensive data understanding is still challenging due to the complexity of various datasets with numerous types and complex metadata.
We, therefore, ask: \emph{can a system achieve a comprehensive understanding of domains and variations in diverse datasets in a domain-agnostic manner as compared to domain-specific systems, such as CoScientist \cite{Boiko2023AutonomousCR}?} 



    





\mypara{Data Transformation.} Different datasets have unique characteristics, requiring custom transformations and filtering operations \cite{Kang2017NoScopeOD}. Moreover, even within the same dataset, different hypotheses may demand different 
transformations for accurate verification and testing. Without this capability, the potential to conduct a wide range of statistical tests for hypothesis verification would be compromised \cite{bailis2017macrobase}. A simple example of data transformation would be the ability to convert a categorical variable into a one-hot encoding. Further, the following is an example showing \osod{}'s LGM-based programmer performing data transformation in order to derive interaction terms between variables: 

\begin{coloredquotation}
Let's start by adding interaction terms to examine the potential link between time preference and BMI across different demographic groups (…)
\textit{Full example}: \Cref{fig:data_transformation_1}
\end{coloredquotation}

The challenge lies in accommodating the 
abundant diversity of
hypotheses and datasets, 
each requiring highly customized transformations \cite{bowers2004ontology}. The ability of LGMs to generate code for such domain-specific data \cite{Sharma2023AutomaticDT} hints towards a generalized solution; however, the 
difficulty in debugging generated code \cite{Vaithilingam2022ExpectationVE} demands a call to action for building better code generation models.




\mypara{Scale.} Modern scientific exploration often involves large amounts of data, a complex analytics workflow, and a large hypothesis space \cite{elliott2016conceptions}. It is important, thus, for a useful autonomous discovery system to be able to sift through such large datasets efficiently 
while maintaining the state of its several processes and tracking previously conducted analyses.
Without this ability to scale and handle complex workflows, several hypotheses would remain unexplored, and valuable insights left undiscovered.

For longitudinal studies, where it is important to understand how variables evolve over time \cite{Weiss1996OverviewOI}, scalability is particularly crucial in order to handle data over extended time periods.
Furthermore, in very large-scale data scenarios, such as the Cancer Moonshot project\footnote{\url{www.whitehouse.gov/cancermoonshot/}} and the Cancer Genomics Cloud \cite{lau2017cancer}, the discovery system must be able to analyze petabytes of data in complex workflows, all while maintaining a state of the possible hypotheses and variable combinations as well as the explorations conducted thus far. In such scenarios, LGMs must be able to support long-horizon planning and long-context attention. However, LGMs are yet to show significant progress on both counts \cite{Valmeekam2022PlanBenchAE}, a limitation of \osod{} as well, thus highlighting a need for focused research towards these goals.




\subsection{Hypothesis Generation}


\mypara{Connecting Data and Scientific Literature.}
The ability to bridge the provided data and existing scientific literature is important 
in providing an understanding of the hypothesis space grounded by contextual domain knowledge.
This ability to learn from known knowledge may further result in various inter-disciplinary perspectives and insights---a phenomenon often called Swanson Linking \cite{bekhuis2006conceptual}.

For example, to derive novel insights between social background and college graduation \cite{alexander1982social} from the National Longitudinal Surveys, it is imperative to understand previous research on National Longitudinal Surveys to avoid duplication and incorporate verified knowledge from the literature to improve initial hypotheses.  

Linking generated hypotheses to existing knowledge requires accurate retrieval, information extraction, and multi-step reasoning \cite{Wang2023LearningTG}. Further, combining multiple research articles connects back to the original Swanson Linking problem \cite{Swanson1986UndiscoveredPK}. While LGMs have recently been shown to perform well in augmenting citations with relevant context based on a user's history \cite{Chang2023CiteSeeAC}, connecting datasets to scientific literature is an open research problem. By utilizing annotated papers for datasets \cite{Palani2023RelatedlySL}, 
we ask: \emph{can a system learn to combine insights from existing literature and a provided dataset in order to discover novel research gaps?}



\mypara{Formulating initial hypotheses.} 
Scientists prioritize experiments based on academic intuition, empirical evidence, and existing theories. In data-driven discovery, this approach is akin to selecting hypotheses from a vast combinatorial space of variable interactions, often extensive for exhaustive exploration \cite{Agrawal2023ArtificialIA}, to identify dependent and independent variables.


For example, to understand the relationship between \emph{education outcome} and \emph{socioeconomic status}, the system should prioritize investigating how the ``rate of completion of BA degree'' is influenced by socioeconomic indicators, such as accumulated wealth and parents' education, as a plausible hypothesis \cite{alexander1982social}. This is non-trivial because it not only requires the system to have a semantic understanding of the variable space 
but also the ability to prioritize hypotheses based on marginal costs and their scientific importance \cite{Agrawal2023ArtificialIA}. Here, \osod{} performs reasonably well on hypothesis generation: 

\begin{coloredquotation}
\textbf{H1:} Females are more likely to complete a BA degree compared to males. \textbf{H2:} Family size has an impact (...). \textbf{H3:} Higher ability scores on the ASVAB test are positively correlated (...)
\textit{Full example}: \Cref{fig:formulate_initial_hypothesis_multi_step_planning_1}
\end{coloredquotation}

Hypothesis generation can be seen as inductive reasoning \cite{Qiu2023PhenomenalYP} using known evidence by connecting them using entailment-like relations \cite{Dalvi2021ExplainingAW}. While LGMs show good performance on reasoning benchmarks \cite{Hendrycks2020MeasuringMM}, data heterogeneity (e.g., variable names, statistical interactions) and semantics make the reasoning problem harder for LGMs \cite{Lu2023MathVistaEM}---thus, we call for research attention.



\subsection{Planning and Orchestrating Research Pathways}

\mypara{Multi-step planning.}
Data-driven discovery with complex problems and datasets requires a structured approach of breaking down a high-level objective into manageable sub-tasks, 
enabling the systematic exploration of the data and hypothesis landscape. This can be considered equivalent to planning \cite{lecun2022path}. 
Prioritized hypothesis search with planning involves \emph{states} -- the intermediate correlations found from data (sub-hypotheses), and \emph{operators} -- the statistical tools and literature to combine verified states (here, sub-hypotheses). Multi-step, iterative planning, thus, comprehensively facilitates the search for
scientific discoveries. 

Research planning involves incorporating known or novel research pathways, such as the order of analyses or the methods used, and they vary depending on the research goal of the exploration. It can be challenging to choose between a standardized or pre-defined flow as compared to a dynamic plan depending on the realized intermediate states of the planning. 
Though LGMs as planners are often faulty \cite{Valmeekam2022PlanBenchAE}, planning within the data hypothesis space presents a fertile ground to systematically benchmark LGMs and improve their abilities. 

For example, analyzing the relationship between college education and socio-economic status from National Longitudinal Surveys \cite{alexander1982social}, the system 
generates the following plan:

\begin{coloredquotation}
\textbf{I.} Understand the data (...) \textbf{II.} Generate initial hypotheses (...) \textbf{III.} Explore combinations of dependent variables (...) \textbf{IV.} Call the ``\texttt{run\_logistic\_regression}'' function (...) \textbf{V.} Repeat step IV for other combinations of dependent variables (...) \textbf{VI.} Document the findings (...) \textbf{VII.} Seek clarity where required (...).
\textit{Full example: }\Cref{fig:formulate_initial_hypothesis_multi_step_planning_1}
\end{coloredquotation}
While the ability to decompose abstract plans into executable sub-plans is heavily explored in coding and symbolic reasoning \cite{Khot2022DecomposedPA}, \osod{} presents a strong base case to improve the efficacy of planning by incorporating dynamic 
strategies that account for
search uncertainties. 

\mypara{Exploration vs. exploitation.} The debate concerning whether exploration should be goal-oriented or randomized is crucial in making novel discoveries \cite{Agarwal2023BringYO}. This applies directly to data-driven discovery, where variable selection 
by the planner directly impacts what subset of the hypothesis space is considered for search. Thus, this exploration-exploitation trade-off is a key factor in shaping the makeup of the final outcome \cite{foster2003serendipity}. 

LGM-based planners, including \osod{}, prefer direct, goal-oriented variables, e.g., preferring \emph{parents' wealth} towards \emph{success in college education}, while de-prioritizing more implicit variables related to urban planning (e.g., \emph{location of schools}). However, while exploration with intrinsic motivators could lead to novel outcomes, it can also sometimes result in false positives \cite{Oudeyer2008HowCW}. How contexts, domains, and the hypothesis space influence the tradeoff between exploration and exploitation remains an open question, which, we argue, is worth considerable research focus \cite{Majumder2022AchievingCG, Burda2018ExplorationBR}. 



\subsection{Hypothesis Evaluation}

\mypara{Hypothesis Verification.} The practical possibility of programmatically verifying a set of hypotheses is a unique feature in data-driven discovery. This encompasses both the proper execution of code as well as the capacity to utilize the appropriate statistical methods and techniques aligned with the high-level research objective \cite{Cai2023LargeLM}.

The verification of hypotheses can involve (1) the use of tools and (2) code generation. 
Tools represent a pre-defined set of structured functions, which may be invoked via function-calling by LGMs along with relevant arguments \cite{Pelrine2023ExploitingNG}. 
Code generation, on the other hand, is often unconstrained and can optionally be combined with external tests \cite{Schfer2023AnEE} and methods such as self-refine \cite{Madaan2023SelfRefineIR} in order to minimize hallucination and execution failure.

For example, to verify the hypotheses proposed by the planner, we show \osod{}'s use of independent t-tests to uncover the impact of wealth distributions in two groups on their incarceration probability \cite{Zaw2016RaceWA}.

\begin{minted}[frame=lines,
baselinestretch=1,
bgcolor=Box3Color,
fontsize=\footnotesize, breaklines, breaksymbolleft={}, breaksymbolright={}]{python}
from scipy import stats
# Perform independent t-tests for the wealth variables across the two groups
test_results_1985 = stats.ttest_ind(df[df['ever_jailed'] == 0] ['composite_wealth_1985'],\n df[df['ever_jailed'] == 1] ['composite_wealth_1985'],\n equal_var=False)(...)
\end{minted}
\vspace{-2.5em}
\begin{coloredquotation}
    The results of the independent t-tests for the wealth variables across the two groups (those with and without a criminal record) for the years 1985, 1990, and 1996: (...) T-statistic: 9.7794 (...)
    \textit{Full Example: }\Cref{fig:hypothesis_verification_1}
\end{coloredquotation}

An ideal system must conduct statistical tests (e.g., correlation, regression, multivariate analyses, t-tests or ANOVA for hypothesis testing, etc.), consume execution results, perform analysis to either conclude or re-plan \cite{Prasad2023ADaPTAD} and support usage of domain-specific evaluation toolkits, such as clinical trials \cite{Rotolo2018surrosurvAR}
and climate change \cite{Hoffmann2021APP}.


The complexity of this task arises from the need to support a plethora of analysis tools (see \Cref{fig:toolbench}) on diverse datasets through unconstrained code generation. Robust verification, further, must be able to analyze execution output and recover from failed initial generation \cite{Ellis2020DreamCoderGG}.
Verification of program output can be enhanced plots, sub-codes, and numerical analyses, yet despite success in math reasoning \cite{Cobbe2021TrainingVT}, LGMs lack multi-modal symbolic understanding \cite{Lu2023MathVistaEM}, calling to action the need for improved data experts in systems like \osod{}.




\mypara{Continual Learning.}  Data-driven discovery is an evolving process. With each stage, from hypothesis generation to evaluation, the system collects new insights and successful (or failed) research flows. The system, thus, requires an adaptive learning approach to integrate and understand the changing context and update its understanding of the dataset \cite{Majumder2023CLINAC, Shinn2023ReflexionLA} over time.

For example, execution errors while running generated code or failed research pathways provide opportunities for self-refinement and possibly integrating learning into the next instances for more fail-proof planning and execution. Continual learning for data-driven discovery opens up research questions regarding the process of online learning \cite{Majumder2023CLINAC, Wang2023VoyagerAO} involving LGMs and avenues to collect supervision signals for continual fine-tuning \cite{Lin2022OnCM}. We argue that how LGMs adapt to novel tools and code at inference time is still an open question and remains critical to data-driven discovery.



\subsection{Measurement of Progress}
\mypara{Measuring intermediate progress.} Unit tests benchmark intermediate progress in software engineering \cite{Lukasczyk2022PynguinAU}. While a parallel does not exist in ML research, data-driven discovery presents quantitative opportunities to develop internal robust benchmarks for progress evaluation---a property non-existent in almost every discovery system, including \osod{}. Akin to FunSearch in \cite{RomeraParedes2023MathematicalDF}, we propose to generate a synthetic benchmark with planted hypotheses that are compositionally verifiable for internal evaluation. The infinitely large space of data-generating functions is potent for exploring such data-generation strategies for robust evaluation.

\mypara{Accommodating human feedback.} Autonomous systems can often get stuck, fall into loops, or fail in other unexpected ways.
Human feedback corrects errors, prevents unintended paths, and provides necessary interventions ensuring that desired objectives are met. \osod{} often 
deviates when fully unsupervised. In the following example, the system focuses on removing multicollinearity despite having a different objective of demographic analysis and having just removed multicollinearity. A user intervention was, thus, necessary.

\begin{coloredquotation}
\textbf{User:} Do not investigate multicollinearity issues. Instead, identify any unique insights or challenges faced by different demographic groups. (...)
\textit{Full example:} \Cref{fig:human_feedback_1}
\end{coloredquotation}

Despite high degree of natural language fluency, LGM-based systems are often not very proactive. It is desirable for these systems to possess a mixed-initiative ability, thus, optimizing the frequency of asking for human feedback and input \cite{Majumder2021AskWM}. Exploring user involvement in the decision-making process raise two questions: (1) \emph{Can we achieve an ideal outcome by enabling users to provide input for tasks like setting low-level objectives or summarizing insights?} (2) \emph{How can we implement effective user intervention during errors or loops to guide the exploration when the system deviates}, as raised in \cite{Lahiri2022InteractiveCG}?




\subsection{Knowledge Integration}

\mypara{Interdisciplinary Knowledge Integration.} 
Integrating interdisciplinary knowledge in data-driven discovery enables the interconnection of diverse domains with the high-level research objective, uncovering nuanced associations and insights often overlooked in a single-domain analysis. The challenge lies in internalizing the complexities of different disciplines and recognizing implicit connections, similar to link prediction \cite{Trouillon2016ComplexEF}. 

For example, while exploring time preference on BMI \cite{smith2005time}, it could be insightful to assess the role of economic pressure on health outcomes, using cultural anthropology to gauge spending habits, considering psychological factors to understand spending patterns, and proposing strategies for public health intervention and effective urban planning---partially achieved by \osod{}.


\mypara{Knowledge Frontiers Support.} 
Knowledge frontiers represent cutting edge scientific exploration and drive groundbreaking discoveries in fields like Machine Learning, gene editing, robotics, and renewable energy \cite{hassabis}. Enhancing data-driven discovery systems by extending exploration, integrating new methods, and collecting more data can facilitate the investigation of novel scientific domains.



To simulate a knowledge frontier, we accessed a popular language agent repository, 
Reflexion \cite{Shinn2023ReflexionLA},
and modified the experiment design following \citet{Majumder2023CLINAC}. The new experimental data was fed to \osod{}, which resulted in the following concrete analysis:
\begin{coloredquotation}
Tasks that are more conceptual or require an understanding of complex systems (e.g., genetics, life stages) seem to be areas where the agent can learn and improve. In contrast, tasks that may involve more practical or hands-on activities (e.g., chemistry mixing, freezing) appear to be more challenging for the agent. (...)
\textit{Full example:} \Cref{fig:knowledge_frontier_1}
\end{coloredquotation}

We seek to obtain emergent behaviors from curiosity-driven exploration and back-linking to knowledge frontiers \cite{Groth2021IsCA}. We raise an open question to automatically search or generate novel datasets \cite{Brickley2019GoogleDS} and conduct novel exploration with user moderation, leading to data-driven scientific discovery. 



\subsection{Research Ethics and Fairness}
\mypara{Reproducible Results.}  
Reproducibility stands as cornerstone of the scientific process \cite{Cao2023TheRO}.
However, persistent challenges in achieving reproducibility across disciplines call for innovative solutions \cite{Magnusson2023ReproducibilityIN}. Complexity arises from the variations in research environments, methodologies, and resource limitations. These factors impede validation and replication of research findings, a phenomenon often evident in fields such as social science, economics, psychology, and biomedicine \cite{Camerer2018EvaluatingTR, OpenScienceCollaboration2015, Fanelli2018OpinionIS}.

For example, \emph{The Reproducibility Project: Psychology} replicated 100 psychology studies and 
found only 36\% of replications to yield significant results, prompting increased awareness and initiatives to enhance reproducibility across scientific disciplines \cite{OpenScienceCollaboration2015}. The ideal discovery system should ensure that the undertaken research pathways are reproducible. \osod{} shows a proof-of-concept for automated, reproducible experiments. However, it can be extended towards automatic documentation and code release, thus further improving transparency. 

\mypara{$p$-hacking Proof.}
Manipulating data or analyses to find false significance undermines the scientific process, leading to unreliable findings and subsequent slowdown of progress. For an automated discovery system, this presents a particularly challenging concern and one that can affect it's trustworthiness 
\cite{Wasserstein2016TheAS}.
For example, consider a scenario where an automated discovery system explores a large dataset to find potential relationships. $p$-hacking might involve tweaking variables or testing multiple hypotheses until a significant result is found \cite{Dunn1961MultipleCA}. The data-driven discovery opens up the unique case of evaluating a significant number of hypotheses at the same time, presenting opportunities for unintentional $p$-hacking.
With a large hypotheses space, there is more chance for accidental findings. An ideal data-driven discovery system must perform tests to counter false discoveries \cite{Korthauer2018APG} to keep the false discovery rate as low as possible.


\section{Limitations of Data-driven Discovery}
\mypara{Hallucinations.} LGM-powered data-discovery struggles with output hallucinations, exacerbated by memorization and superposition issues \cite{elhage2022superposition} -- most susceptible being hypothesis generation, planning, and output comprehension. This undermines the benefits of automation, necessitating external verification and user moderation.



\mypara{Cost at scale.} In high-throughput fields (e.g., computational biology), it is common to test millions of hypotheses \cite{korthauerBiochem}. Extensive reliance on these systems for orchestrating experiments can then incur significant computational costs---highlighting the need for integrated cost-benefit analyses into the discovery systems \cite{Agrawal2023ArtificialIA} using, for instance, predictive hazard functions.

\mypara{Policy misuse.} The autonomous discovery system is always at risk of misuse by bad actors to produce a substantial volume of dubious research to fit a particular agenda \cite{heaven2022meta}. For certain disciplines like social science and economics, this could potentially impact policy-making institutions and result in sub-optimal policies and decision-making \cite{Groh2022HumanDO}.

\mypara{Legal Implications.} Autonomous hypothesis generation and verification, supported by datasets, raise legal challenges around intellectual property rights and authorship \cite{CallisonBurch_2023} and liability in decision-making processes involving these systems \cite{ai2response2023}. Defining responsibilities and establishing institutional, legal frameworks to navigate potential suboptimal policies are essential aspects of addressing this challenge. 


\mypara{Underlying Bias.} An inherent challenge with the data-driven discovery system involves the potential percolation of bias originating from dual sources---the underlying dataset \cite{Caliskan2016SemanticsDA} and the LGMs \cite{Feng2023FromPD}. This introduces the risk of generating hypotheses that reflect and perpetuate existing biases present in the data source being utilized, potentially leading to skewed or unfair insights.

\section{Survey on Related Systems}

\mypara{End-to-end Data-driven Discovery} 
Most previous autonomous data-driven discovery systems, such as Bacon \citep{Langley1981DataDrivenDO, Langley1984TheSF, Langley1983RediscoveringCW} severely lacked the requisite computational power, restricting their scope with limited discovery of data-driven knowledge. A recent system, CoScientist \citet{Boiko2023AutonomousCR}, uses LGMs to automate some parts of the workflow, however, still requires substantial human intervention (e.g., wet lab experiments) for hypothesis verification, thus not qualifying as a fully autonomous discovery system. DataLume \cite{Gu2023HowDD} fully automates the code generation for data transformation and hypothesis verification; however, do not have modules to support hypothesis search and orchestrating complex science workflows. \citet{Gil2022TowardsCS, Gil2017TowardsCS, Gil2013UsingSW} as well as Automatic Analysis in WolframAlpha\footnote{\url{https://www.wolframalpha.com/examples/pro-features/data-input}} prototyped various workflows for conducting science in data-driven ways, however, such prototypes never explored the power of LGMs and are only exhibit limited generalizability to datasets and scientific methods. 


\mypara{AutoML}
AutoML is a workflow of automatically building optimal machine learning and predictive models. AutoML tools exist in scientific packages like Scikit \cite{feurer2015efficient} and also in cloud platforms such as Google Cloud Platform\footnote{\url{cloud.google.com/automl}}, Microsoft Azure\footnote{\url{azure.microsoft.com/en-us/products/machine-learning/automatedml/}}, and Amazon Web Services\footnote{\url{aws.amazon.com/machine-learning/automl/}}. Existing AutoML Cloud platform systems mainly focus on black box models, ensuring the models can be served at scale. Despite performing searches over hyperparameter space for optimal model development, these systems cannot comprehend the semantics of the data and hence cannot help with data-driven hypothesis generation, planning, orchestrating research pathways, and knowledge integration. MLAgentBench \cite{Huang2023BenchmarkingLL} can be considered as an evolution of AutoML that performs end-to-end machine learning to benchmark AI research agents. MLAgentBench can plan, evaluate hypotheses, and measure progress, but with a focus on optimizing machine learning models, not on discovering new and novel scientific knowledge.


\mypara{Automated Data Analysis}
Automated Data Analysis tools are primarily focused on exploring data under a user-provided hypothesis or query (e.g., ``plot sales trends for last 12 months'', etc.) and often do not have the capability of searching through the hypotheses space as defined by the data. Tools such as PowerBI\footnote{\url{https://www.microsoft.com/en-us/power-platform/products/power-bi}}, Tableau\footnote{\url{https://www.tableau.com/}}, and Thoughtspot\footnote{\url{https://www.thoughtspot.com/}} can, however, perform multi-step hypothesis verification using inbuilt data transformation and statistical tools, though the interpretation and consumption of such analysis are left to the user. Spreadsheet tools such as Microsoft Excel and Google Sheets are often part of the scientific workflow as the data analysis and show limited ability for an autonomous data analysis framework even after having Python integration or coding support \cite{PYExcel, duetai}. Focus on integrating LGMs into data analysis with known workflows \cite{Perlitz2022nBIIGAN, Chakraborty2024NavigatorAG}  and code-first data analysis \cite{ydata} increased recently---however, these are limited to small-scale tables and lack abilities such as the ability to orchestrate research plans, interpret results, measurement of progress, and knowledge integration.

\section{Conclusion}
We argue that
ongoing ML research on reasoning, planning, code generation, and tool utilization with LGMs can have
a significant influence on advancing and accelerating
data-driven discovery. Such systems can transform domains overwhelmed with vast amounts of data, including but not limited to observational social sciences, medicine, astronomy, biology, climate science, computational science, consumer science, and social media analytics.

We posit that the time is ripe for advancing data-driven discovery, and that integrating LGMs with tools and user feedback can catalyze notable progress in scientific inquiry.
We hope our timely position can increase interest and efforts in developing, debating, and enhancing the vision for an accurate, reliable, and robust system for data-driven discovery. It can help initiate a Cambrian explosion of discovery while promoting speed, reproducibility, and collaboration in scientific research.



\section{Impact Statement}
This position paper presents arguments for a goal to advance the field of science by building end-to-end data-driven discovery systems using ML. There are many potential societal consequences of our proposed direction since it involves using large generative models, some of which we cover in our Limitations section, including policy misuse, legal ramifications, and false discovery. On the positive side, our proposed system can advance the rate of discovery, leading to an improved standard of living and social well-being.

\section*{Acknowledgments}
We sincerely thank Abhijeetsingh Meena, Aryan Prakhar, and Tirth Vora for their engineering and exploration efforts in making \osod{}. We also thank Peter Jansen, David Wadden, Yoav Golberg, and Daniel Weld for their useful comments. We thank Siddharth Sharma and Siddharth Narayanan for their help with proofreading.


\bibliography{example_paper}
\bibliographystyle{icml2024}

\newpage
\appendix
\onecolumn
\appendix




\begin{figure*}
    \centering
    \includegraphics[width=\linewidth]{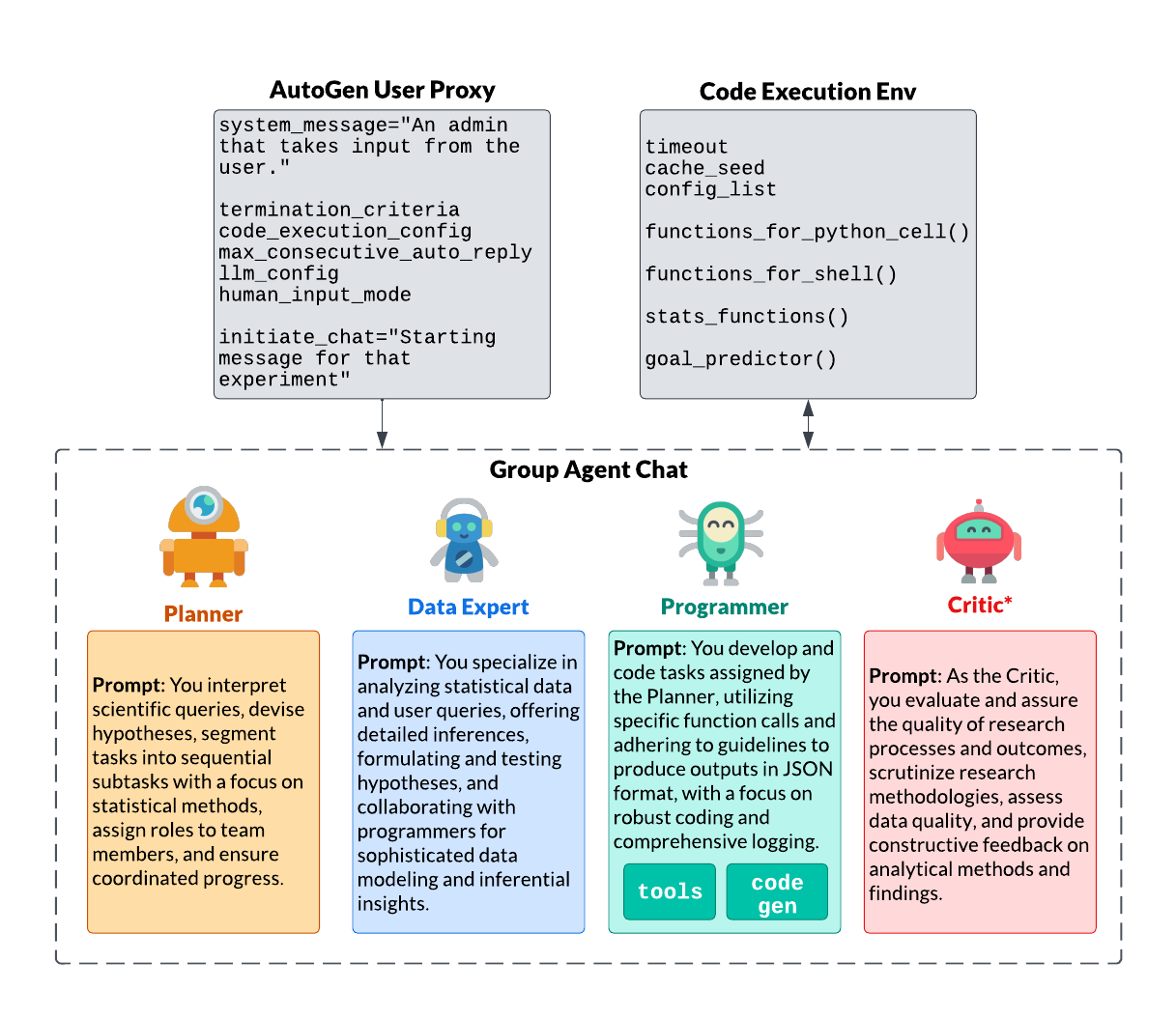}
    \caption{Agent Structure for \osod{}. Group Agent Chat has AutoGen agents that communicate with each other. The User Proxy links the user with the agents to share data, feedback, and goals. Code Execution Environment has access to structured functions and code generation methods that can be called depending on the context.}
    \label{fig:agent-structure}
\end{figure*}


\begin{figure*}
    \centering
    \includegraphics[width=\linewidth]{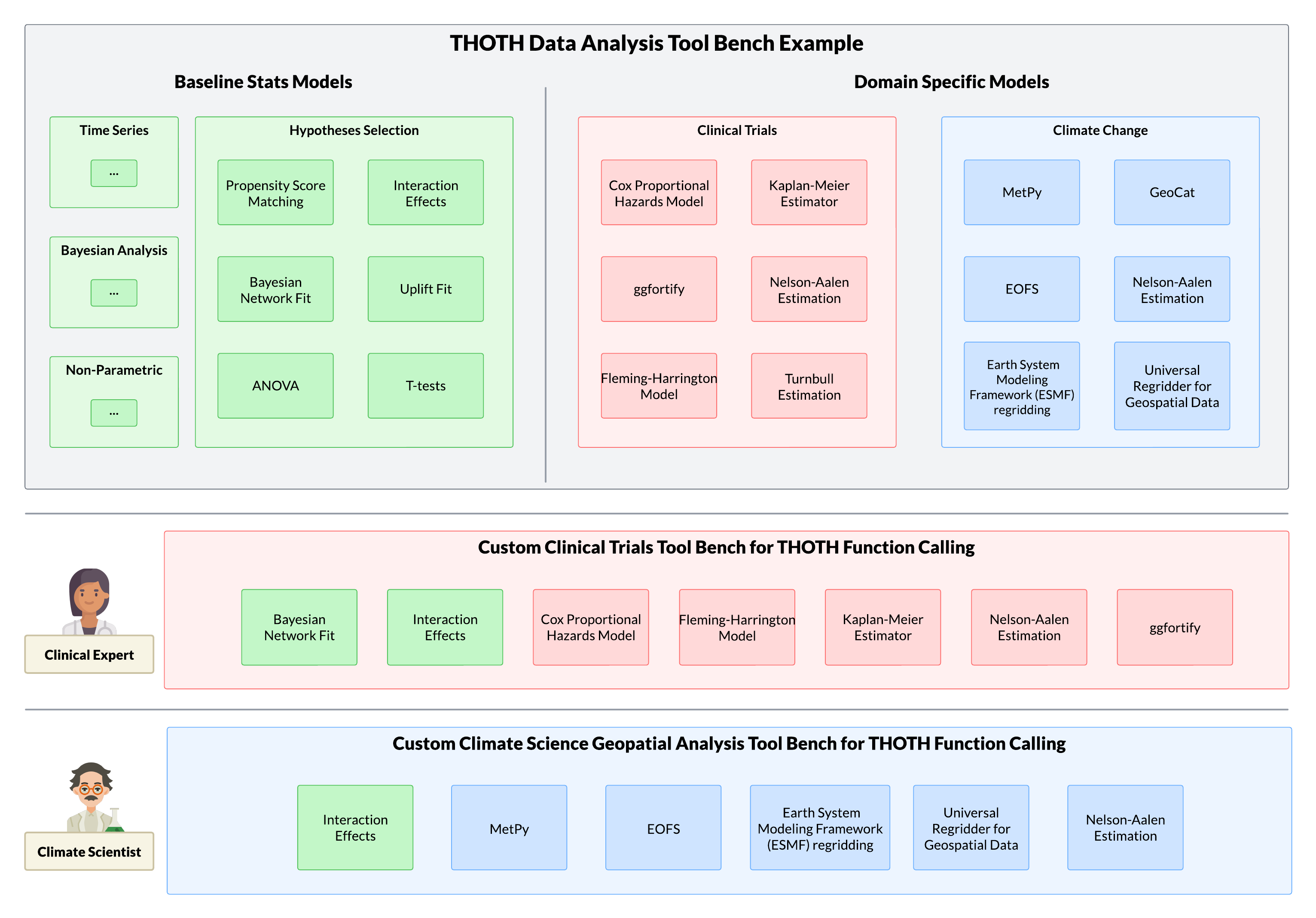}

    \caption{Data Analysis Tool Bench that can be structured inside \osod{} to enable discovery in a wide range of scientific domains.}
    \label{fig:toolbench}
\end{figure*}

\begin{figure*}
    \centering
    \includegraphics[width=\linewidth]
    {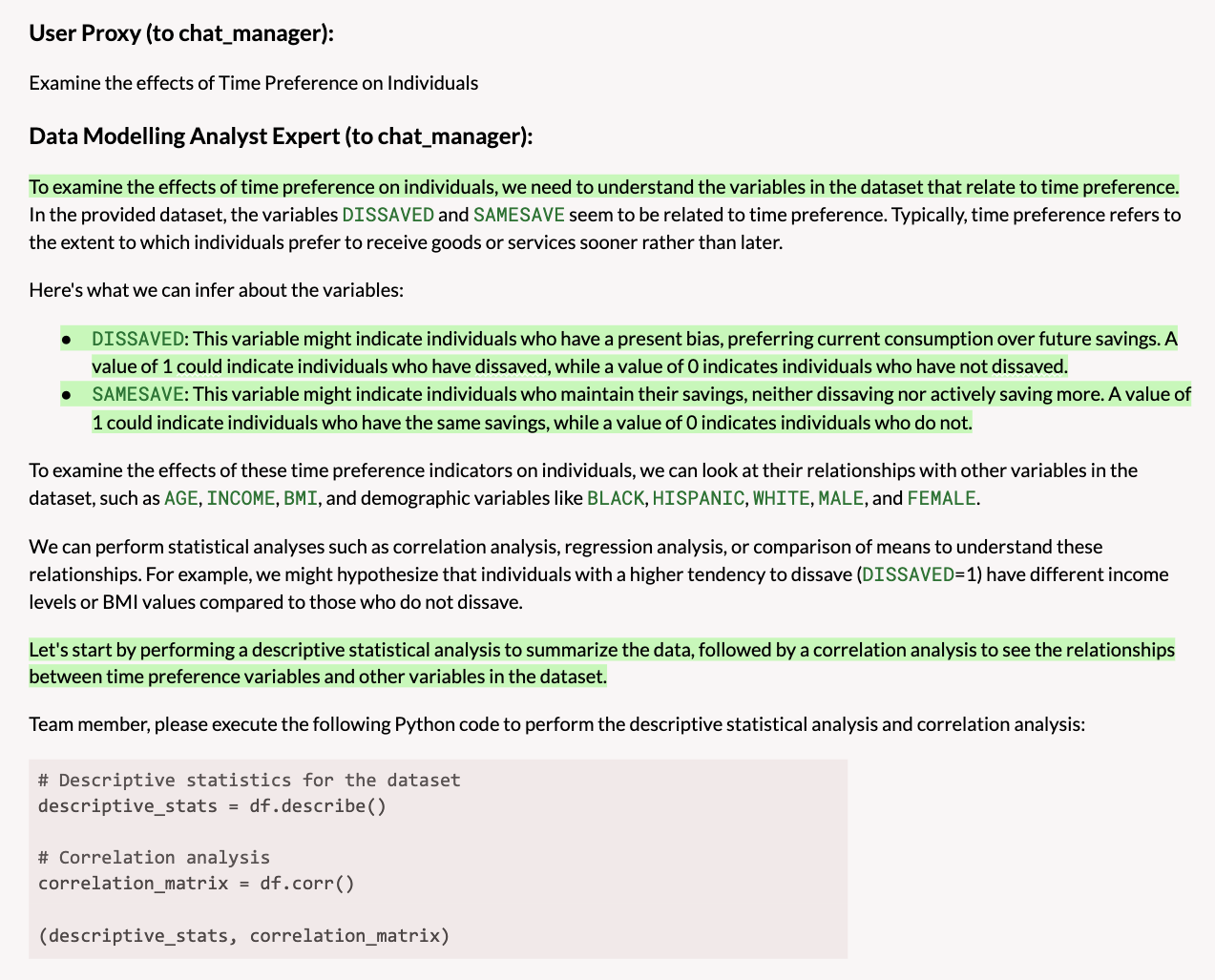}
    \caption{\textbf{Background:} Data from National Longitudinal Survey of Youth along with question on relation between time preference \& BMI was fed into \osod{}; it is a question studied in \cite{smith2005time}. \textbf{This figure:} Data Understanding - In response to a high-level objective, the system demonstrates the need of understanding the variables before initiating statistical analysis. Moreover, it selects relevant Time Preference variables from the data and infers them (highlighted in green).}
    \label{fig:data_understanding_1}
\end{figure*}

\begin{figure*}
    \centering
    \includegraphics[width=0.8\linewidth]
    {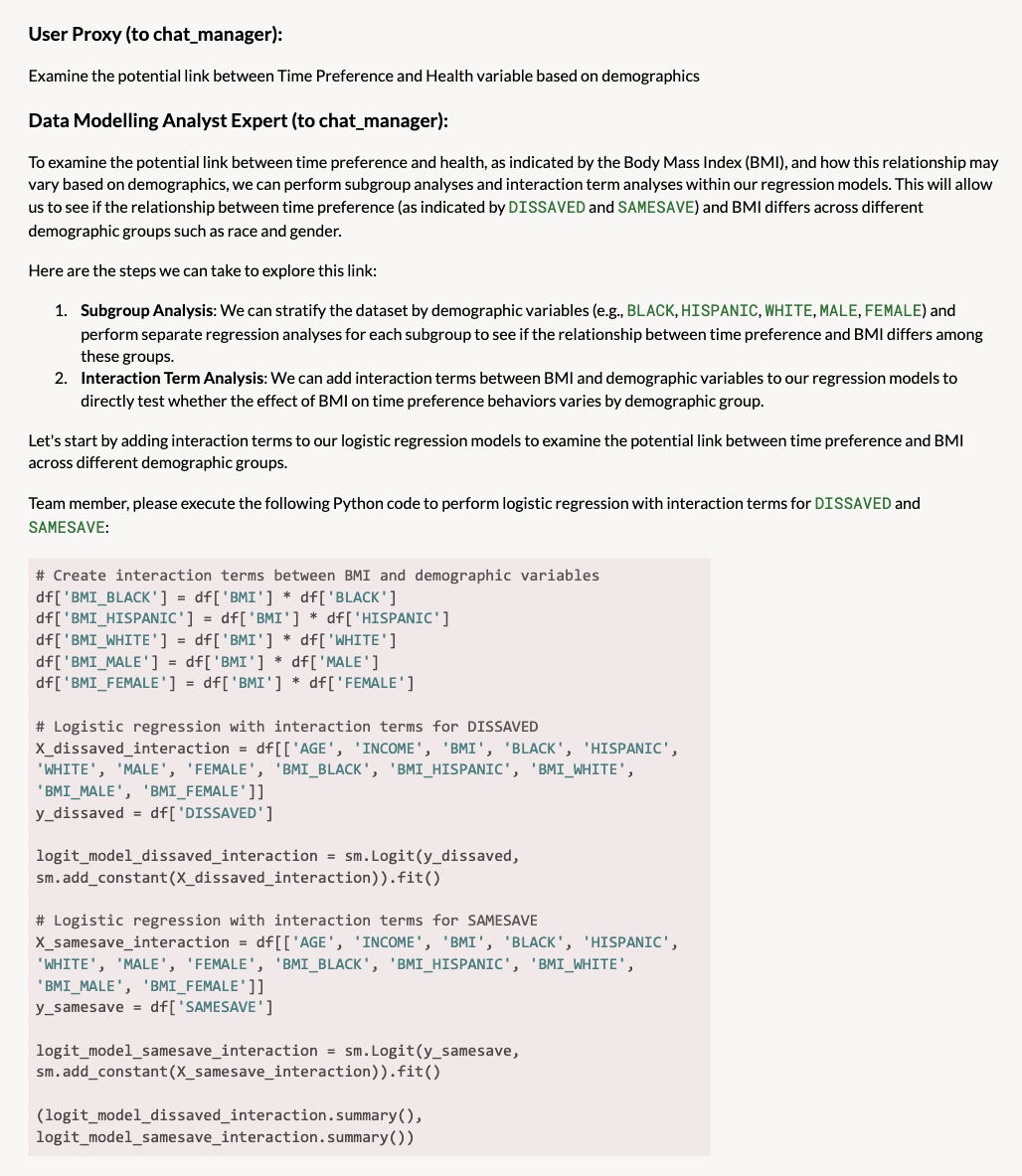}
    \caption{\textbf{Background:} Data from National Longitudinal Survey of Youth along with question on relation between time preference \& BMI was fed into \osod{}; it is a question studied in \cite{smith2005time}. \textbf{This figure:} Data Transformation - The system generated insights from the results of Logistic Regression with L1 regularization. As a response to user input, the system showcases data transformation ability by creating new interaction terms in logistic regression models (as demonstrated in the code snippet), exploring the link between time preference and BMI across diverse demographic groups.}
    \label{fig:data_transformation_1}
\end{figure*}

\begin{figure*}
    \centering
    \includegraphics[width=0.98\linewidth]{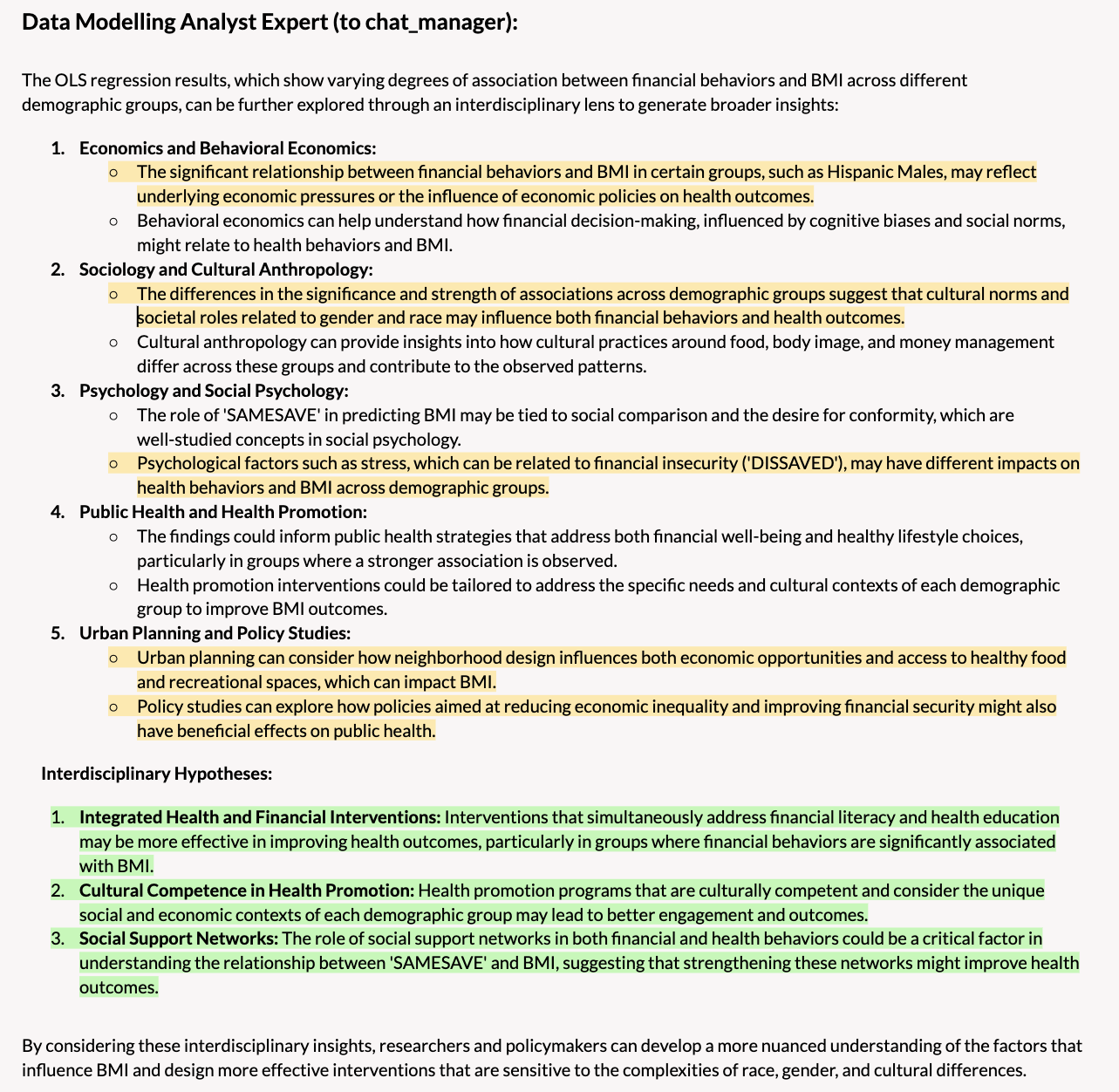}
    \caption{\textbf{Background:} Data from National Longitudinal Survey of Youth along with question on relation between time preference \& BMI was fed into \osod{}; it is a question studied in \cite{smith2005time}. \textbf{This figure:} Interdisciplinary Knowledge Integration - The system extracts insights from BMI data, generates insights (highlighted in yellow) from the lens of different disciplines and integrates them into different interdisciplinary hypotheses for further exploration (highlighted in green).}
    \label{fig:enter-label}
\end{figure*}

\begin{figure*}
    \centering
    \includegraphics[width=0.98\linewidth]{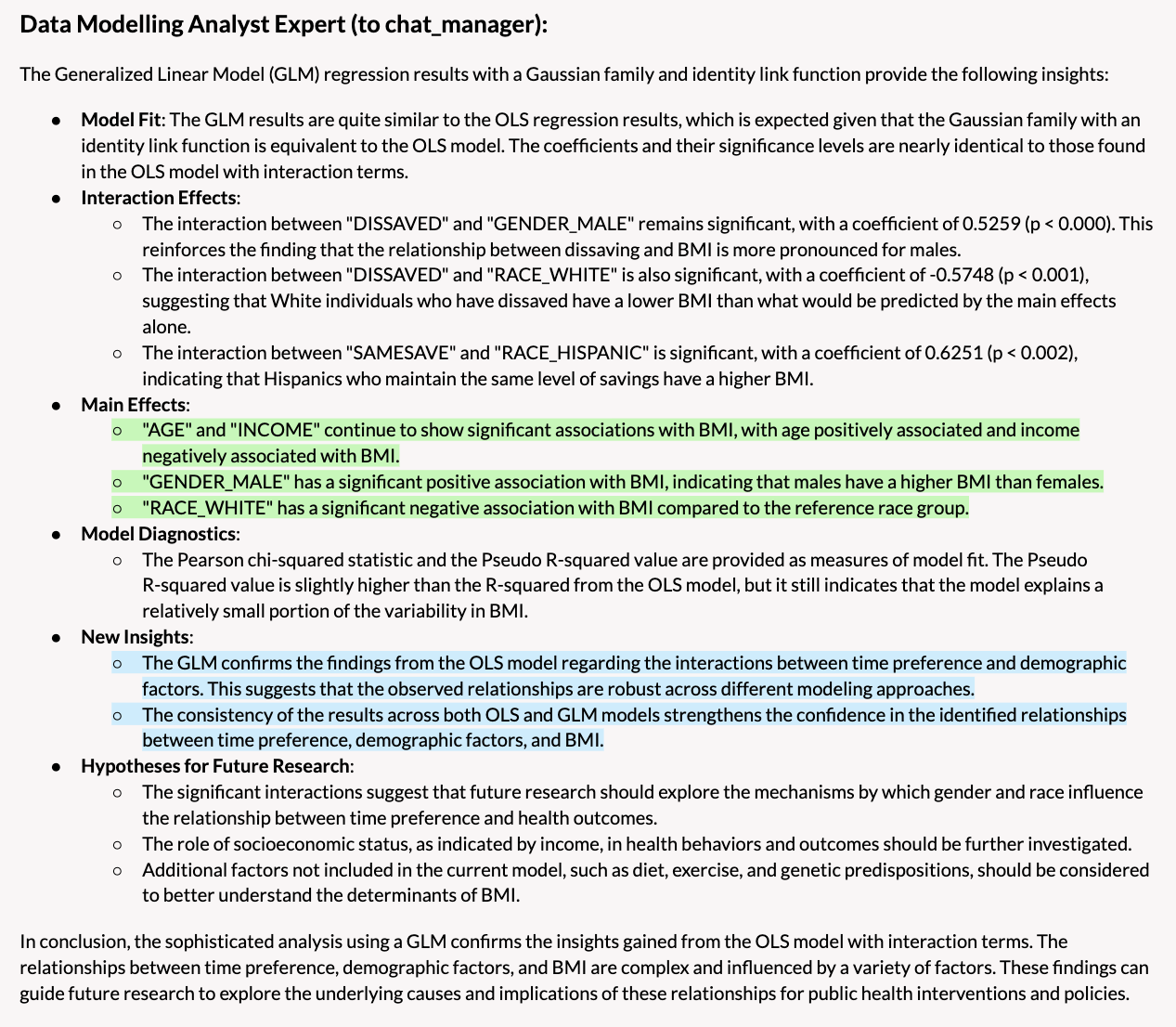}
    \caption{\textbf{Background:} Data from National Longitudinal Survey of Youth along with question on relation between time preference \& BMI was fed into \osod{}; it is a question studied in \cite{smith2005time}. \textbf{This figure:} Hypothesis Verification - When the user prompted to perform sophisticated analysis to uncover new insights, the system generates new insights utilizing the Generalized Linear Model (highlighted in blue) that confirms the results from the previous OLS analysis (highlighted in green).}
    \label{fig:enter-label}
\end{figure*}


\begin{figure*}
    \centering
    \includegraphics[width=0.98\linewidth]
    {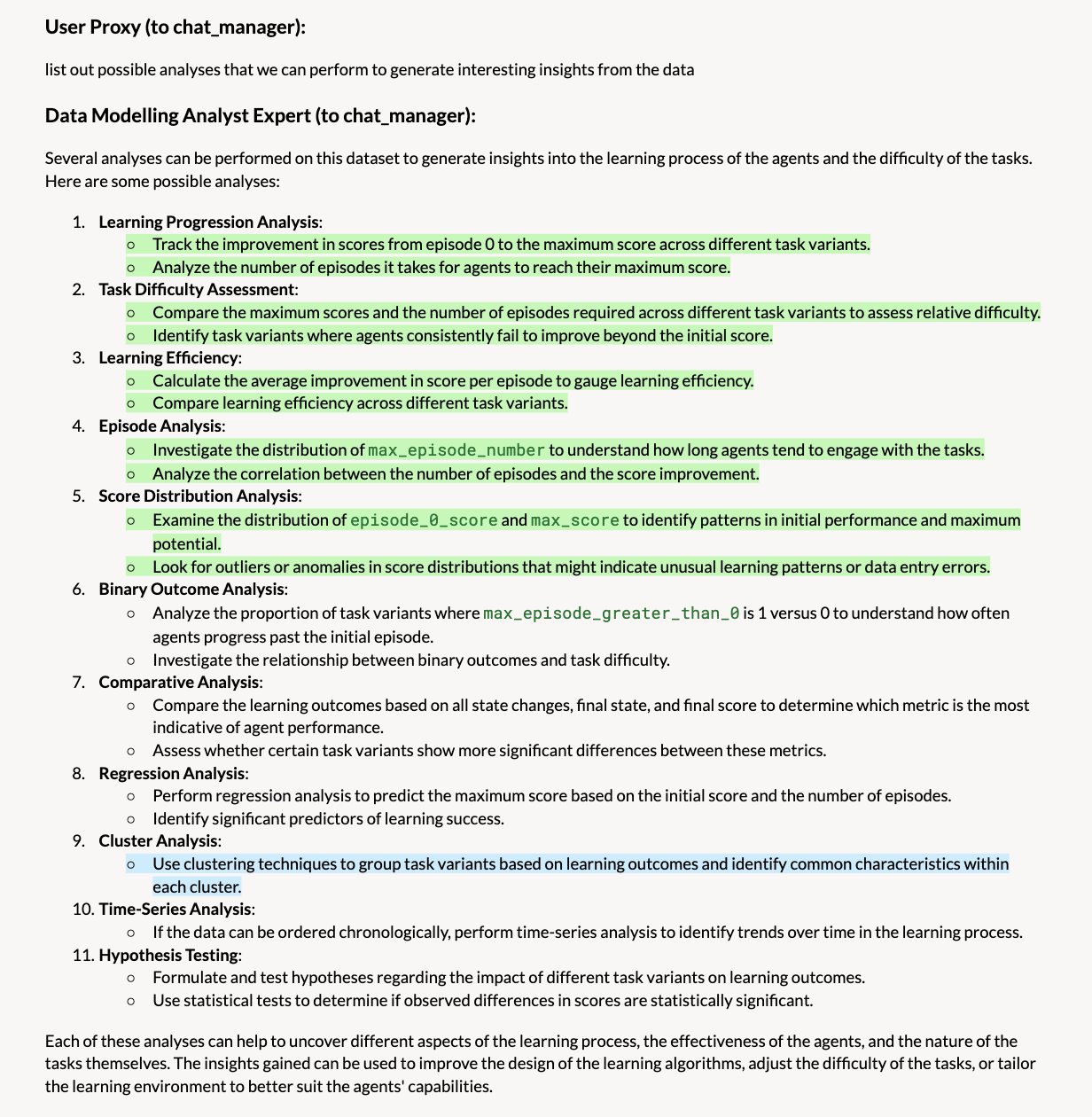}
    \caption{\textbf{Background:} Experimental data of running methods using a popular agent-based repo Reflexion (\url{https://github.com/noahshinn/reflexion}) and \cite{Majumder2023CLINAC} was fed to \osod{}. \textbf{This figure:} Knowledge Frontiers Support - Data Expert suggested interesting list of analyses to find new insights. New analyses (highlighted in green) were created with limited context on the data just based on variable description. Cluster analysis (highlighted in blue) leads to novel insights in the agent literature.}
\end{figure*}

\begin{figure*}
    \centering
    \includegraphics[width=0.98\linewidth]
    {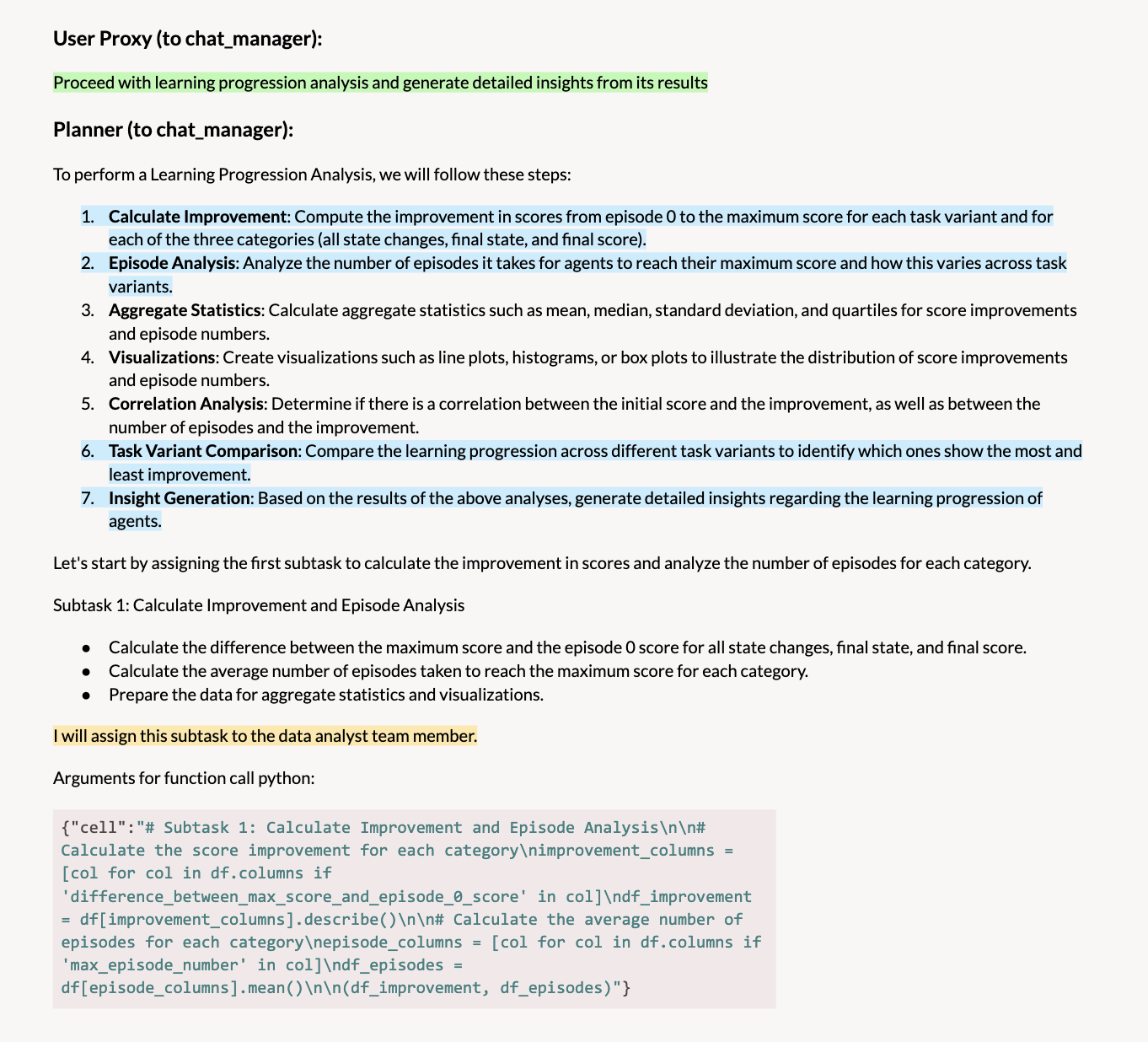}
    \caption{\textbf{Background:} Experimental data of running methods using a popular agent-based repo Reflexion (\url{https://github.com/noahshinn/reflexion}) and \cite{Majumder2023CLINAC} was fed to \osod{}. \textbf{This figure:} Multi-Step Planning - The system understood the variables and carved out the steps that need to be performed to draw interesting insights (highlighted in blue). The Planner created an excellent plan by breaking the objective into subtasks to carry-out learning progression analysis. It then assigned the subtask to a team member (highlighted in yellow). }
\end{figure*}

\begin{figure*}
    \centering
    \includegraphics[width=0.98\linewidth]
    {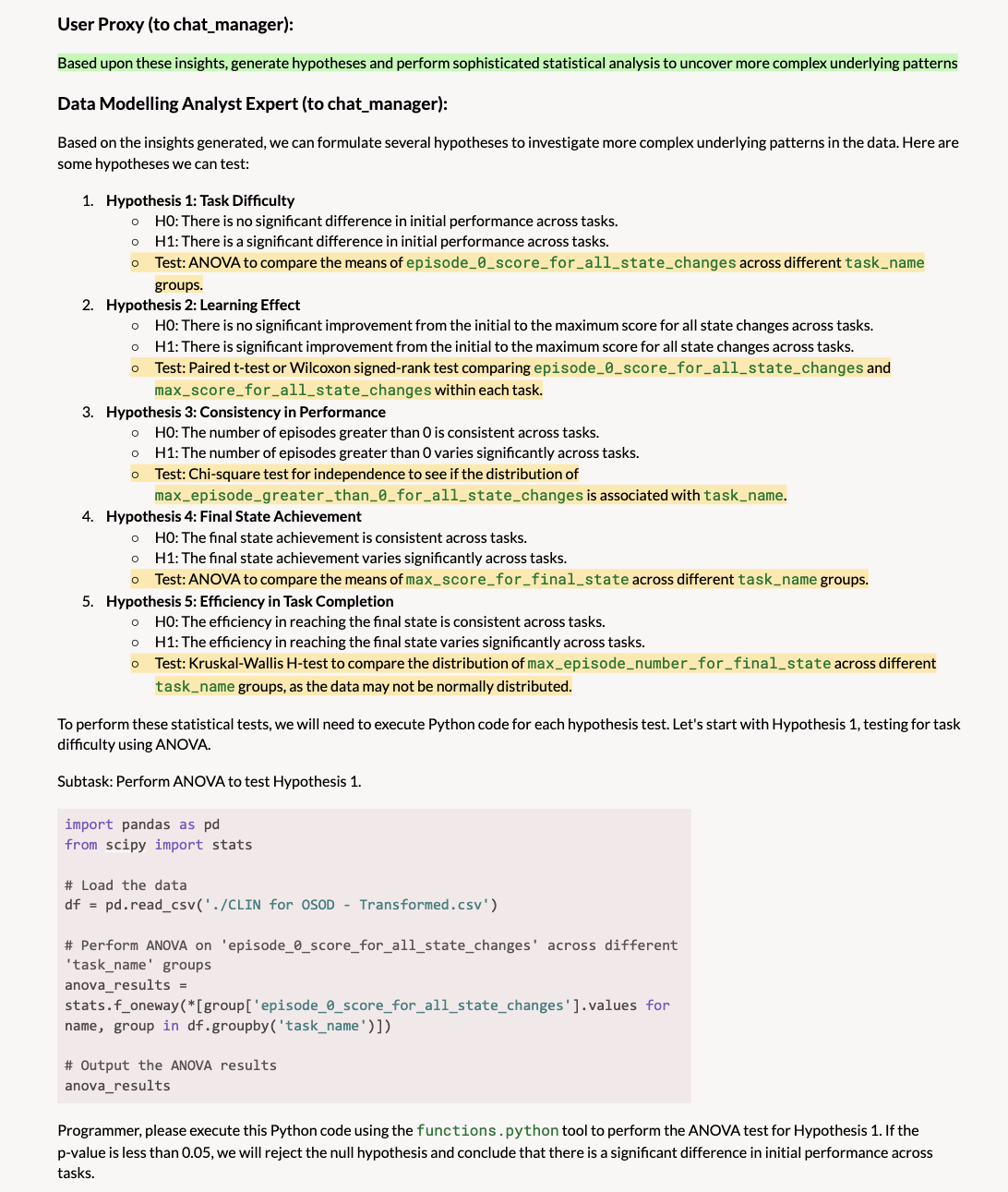}
    \caption{\textbf{Background:} Experimental data of running methods using a popular agent-based repo Reflexion (\url{https://github.com/noahshinn/reflexion}) and \cite{Majumder2023CLINAC} was fed to \osod{}. \textbf{This figure:} Hypothesis Generation - When asked to generate hypotheses and perform sophisticated analysis based on insights (highlighted in green), the system generates testable hypotheses, formulating clear null (H0) and alternative (H1) hypotheses with corresponding statistical tests (highlighted in yellow) for uncovering complex underlying patterns in the data.}
\end{figure*}

\begin{figure*}
    \centering
    \includegraphics[width=\linewidth]{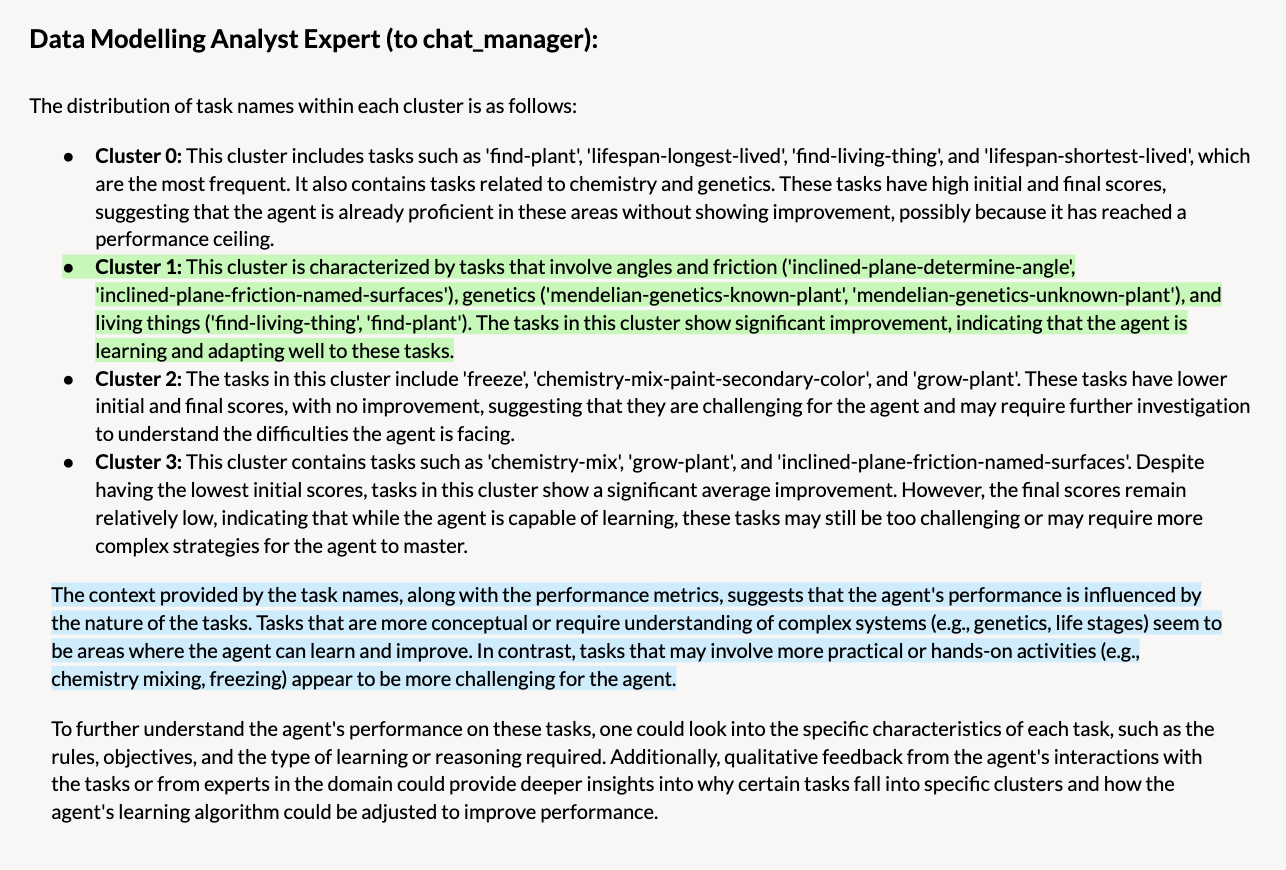}
    \caption{\textbf{Background:} Experimental data of running methods using a popular agent-based repo Reflexion (\url{https://github.com/noahshinn/reflexion}) and \cite{Majumder2023CLINAC} was fed to \osod{}. \textbf{This figure:} Knowledge Frontiers Support - Ability to support and generate new insights on frontiers of knowledge, where novel insights on agents' behavior were generated. The new insights are highlighted in blue.}
    \label{fig:knowledge_frontier_1}
\end{figure*}


\begin{figure*}
    \centering
    \includegraphics[width=0.98\linewidth]{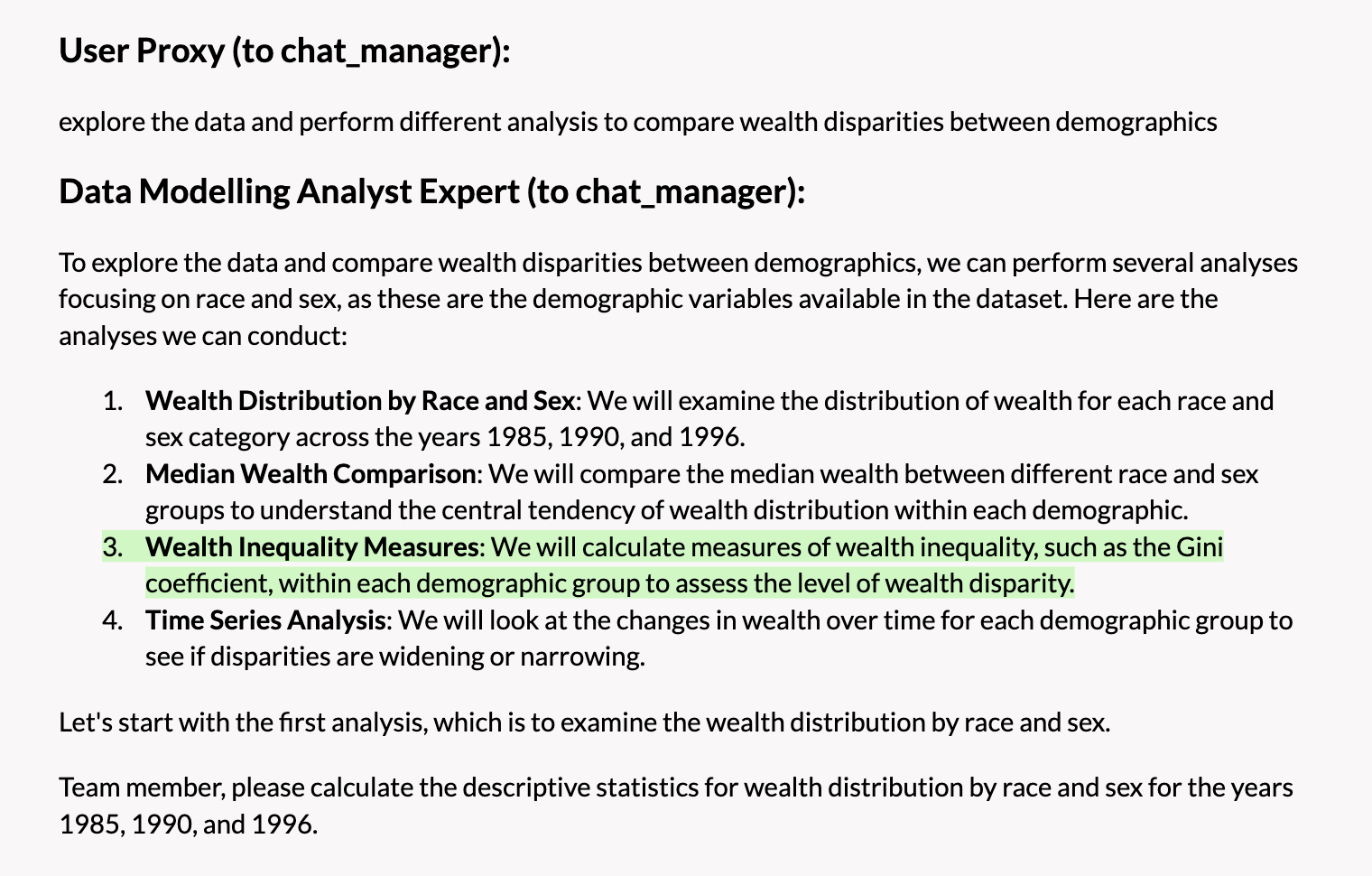}
    \caption{\textbf{Background:} National Longitudinal Survey of Youth data with a question on how incarceration and race affected wealth was fed to \osod{}; it is a question studied in \cite{Zaw2016RaceWA}. \textbf{This figure:} Knowledge Frontiers Support - Despite the original paper talking about wealth analysis post-incarceration and only doing basic statistical analysis over the data, the system was able to suggest new techniques like the application of Gini coefficients - a popular measure used in understanding wealth disparities (highlighted in green).}
    \label{fig:enter-label}
\end{figure*}

\begin{figure*}
    \centering
    \includegraphics[width=0.98\linewidth]{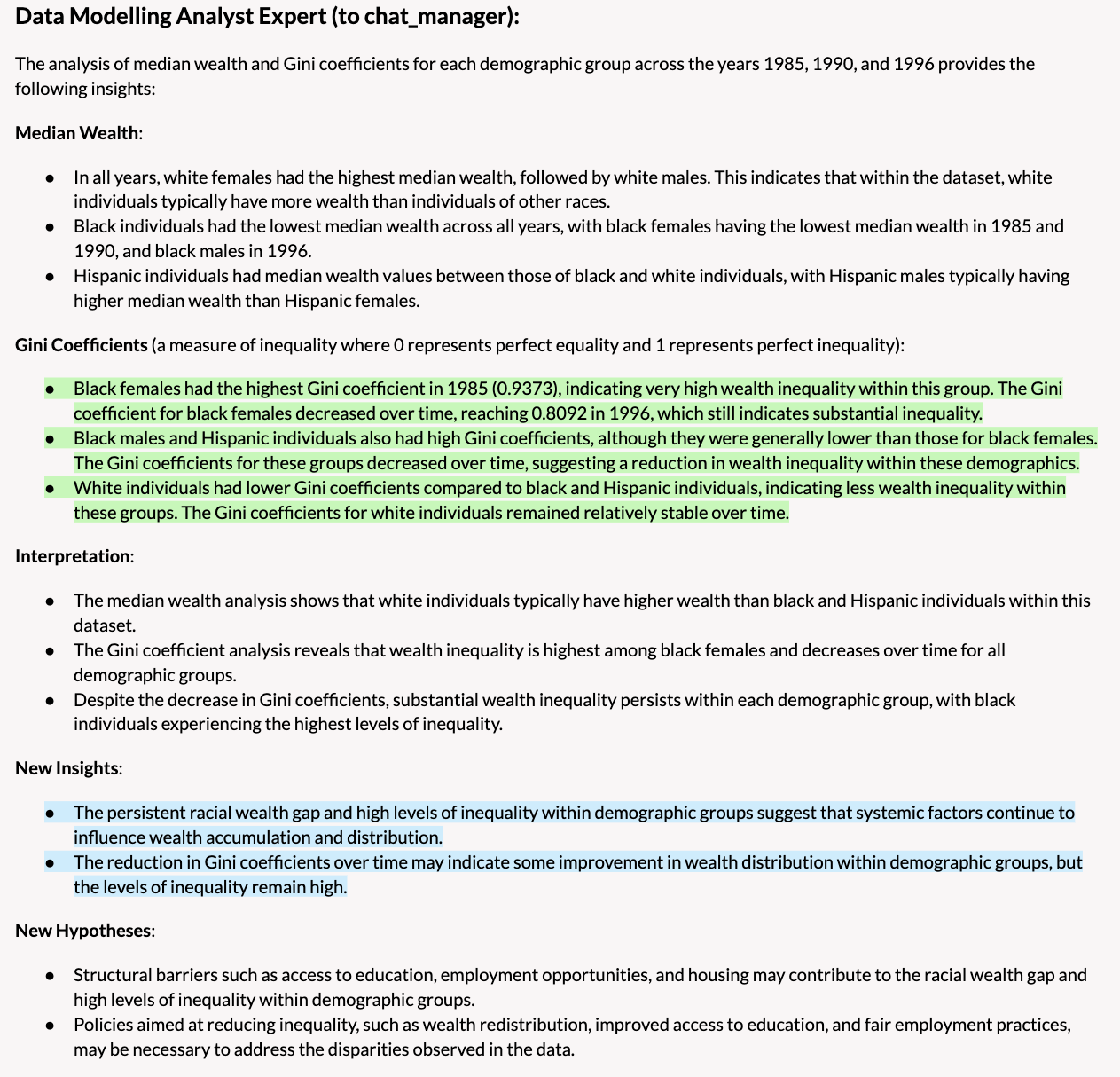}
    \caption{\textbf{Background:} National Longitudinal Survey of Youth data with a question on how incarceration and race affected wealth was fed to \osod{}; it is a question studied in \cite{Zaw2016RaceWA}. \textbf{This figure:} Hypothesis Verification - After calculating Wealth Inequality across demographic groups using gini coefficients, the system interpreted the results (highlighted in green) and generated interesting insights (highlighted in blue).}
    \label{fig:enter-label}
\end{figure*}

\begin{figure*}
    \centering
    \includegraphics[width=0.98\linewidth]
    {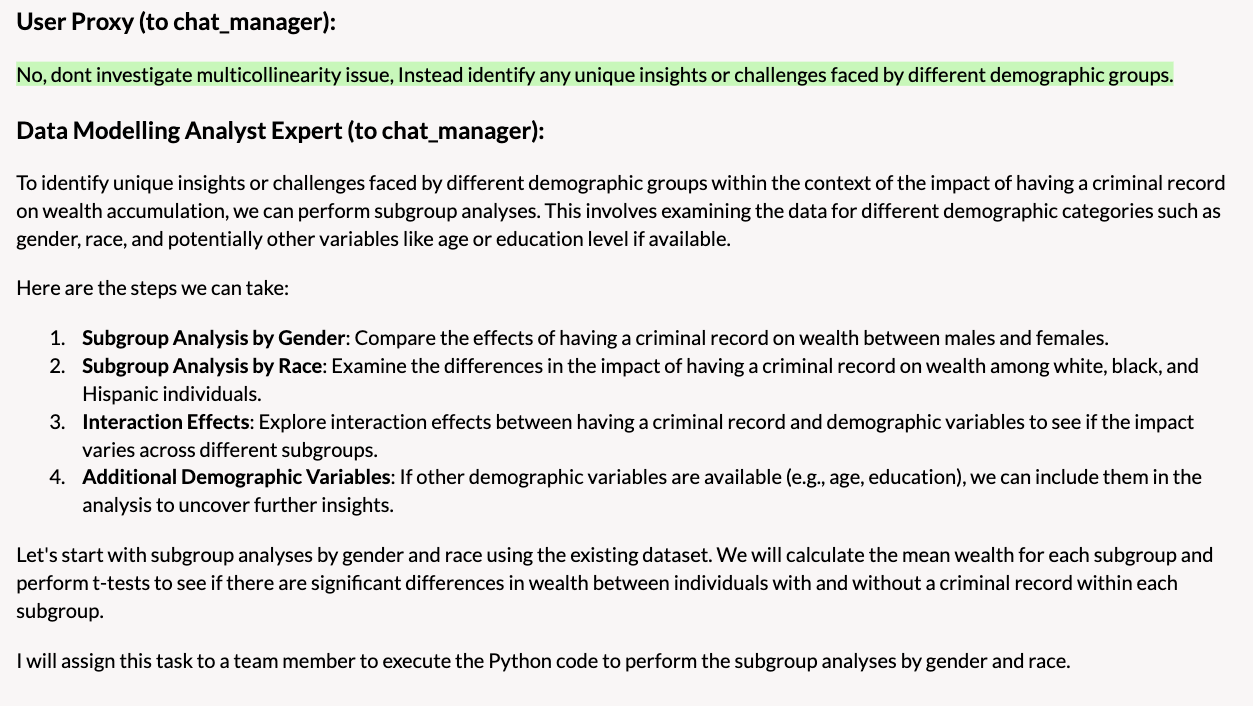}
    \caption{\textbf{Background:} National Longitudinal Survey of Youth data with a question on how incarceration and race affected wealth was fed to \osod{}; it is a question studied in \cite{Zaw2016RaceWA}. \textbf{This figure:} Human Feedback Accommodation - The system performed OLS Regression, suggested the presence of multicollinearity, and removed it using Variation Inflation Factor. Then, the system was set to address multicollinearity again, but user intervention prevented redundancy and redirected it to the objective (highlighted in green).}
    \label{fig:human_feedback_1}
\end{figure*}

\begin{figure*}
    \centering
    \includegraphics[width=\linewidth]
    {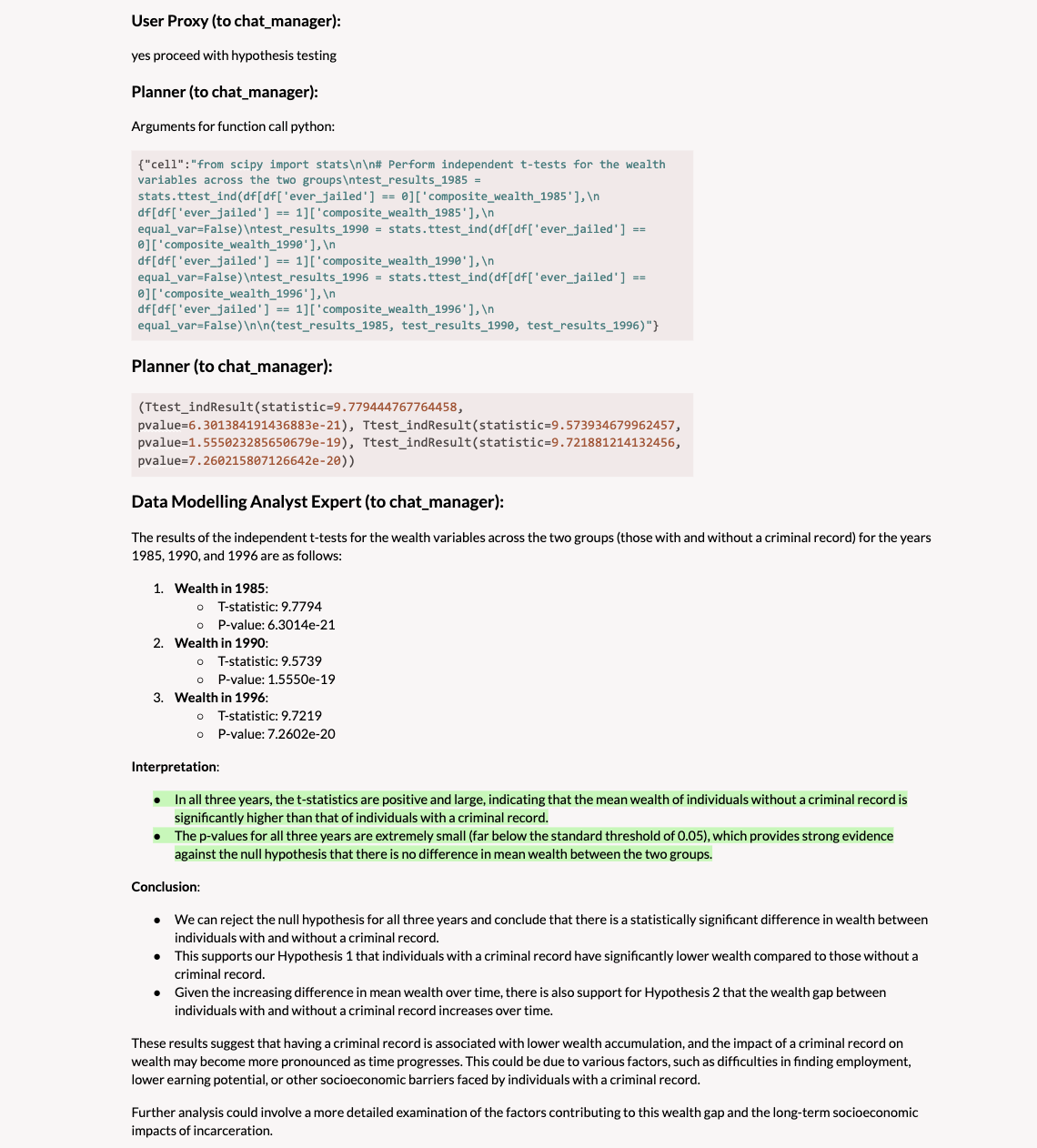}
    \caption{\textbf{Background:} National Longitudinal Survey of Youth data with a question on how incarceration and race affected wealth was fed to \osod{}; it is a question studied in \cite{Zaw2016RaceWA}. \textbf{This figure:} Hypothesis Verification - Following the results of Descriptive Statistics, the Data Expert proposed two hypotheses. When the user prompted the system to perform Hypothesis Testing it verified them by performing T-tests, interpreted them (highlighted in green) and shared the conclusions.}
    \label{fig:hypothesis_verification_1}
\end{figure*}


\begin{figure*}
    \centering
    \includegraphics[width=\linewidth]{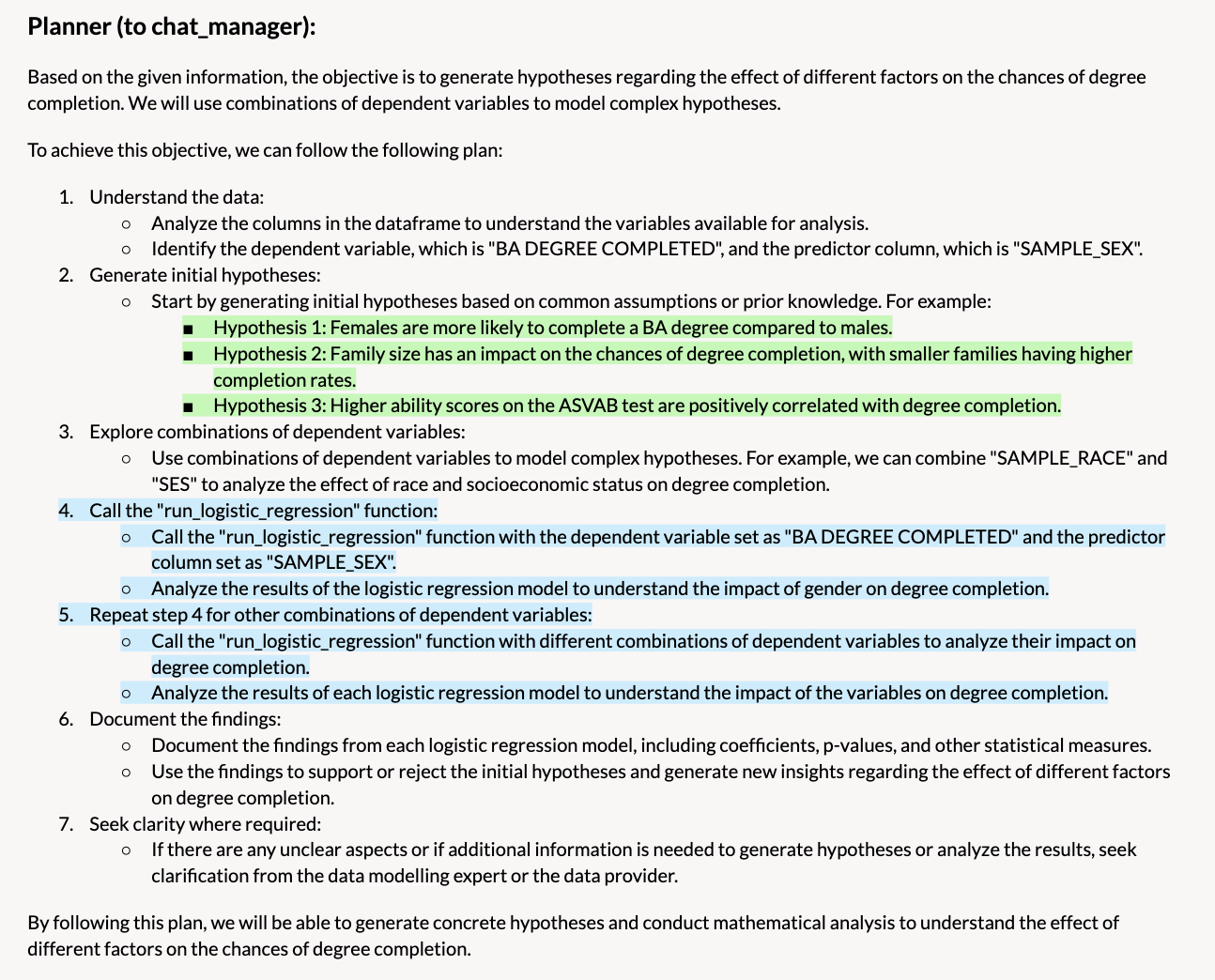}
    \caption{\textbf{Background:} Data from 1979 follow-up wave of the National Longitudinal Survey along with the question on how social background affects degree completion was fed to \osod{}; it is a research covered in \cite{alexander1982social} \textbf{This figure:} Multi-step planning for hypothesis generation - following an initial analysis of data statistics and considering the specified goal, the system formulates a detailed plan to guide the hypothesis generation process. It generates a possible list of hypotheses (highlighted in green) and the core experimental loop (highlighted in blue.)}
    \label{fig:formulate_initial_hypothesis_multi_step_planning_1}
\end{figure*}

\begin{figure*}
    \centering
    \includegraphics[width=\linewidth]{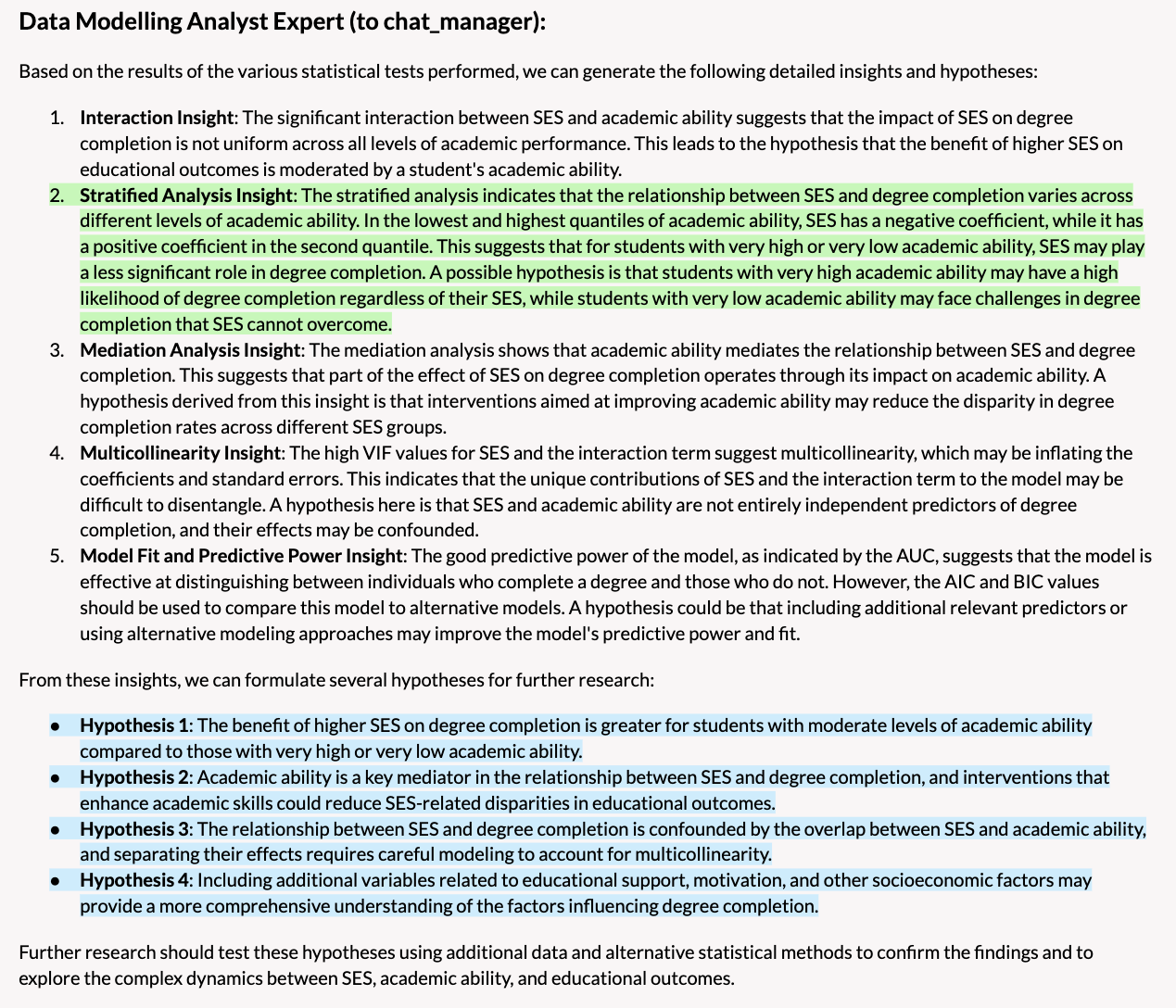}
    \caption{\textbf{Background:} Data from 1979 follow-up wave of the National Longitudinal Survey along with the question on how social background affects degree completion was fed to \osod{}; it is a research covered in \cite{alexander1982social} \textbf{This figure:} Hypothesis Generation - The system conducts new experiments to comprehend SES data and generates hypotheses. Building upon initial statistical tests, the model delves deeply into proposing and conducting more sophisticated experiments (highlighted in green), subsequently formulating several hypotheses for further analysis (highlighted in blue).}
    \label{fig:enter-label}
\end{figure*}

\begin{figure*}
    \centering
    \includegraphics[width=0.95\linewidth]{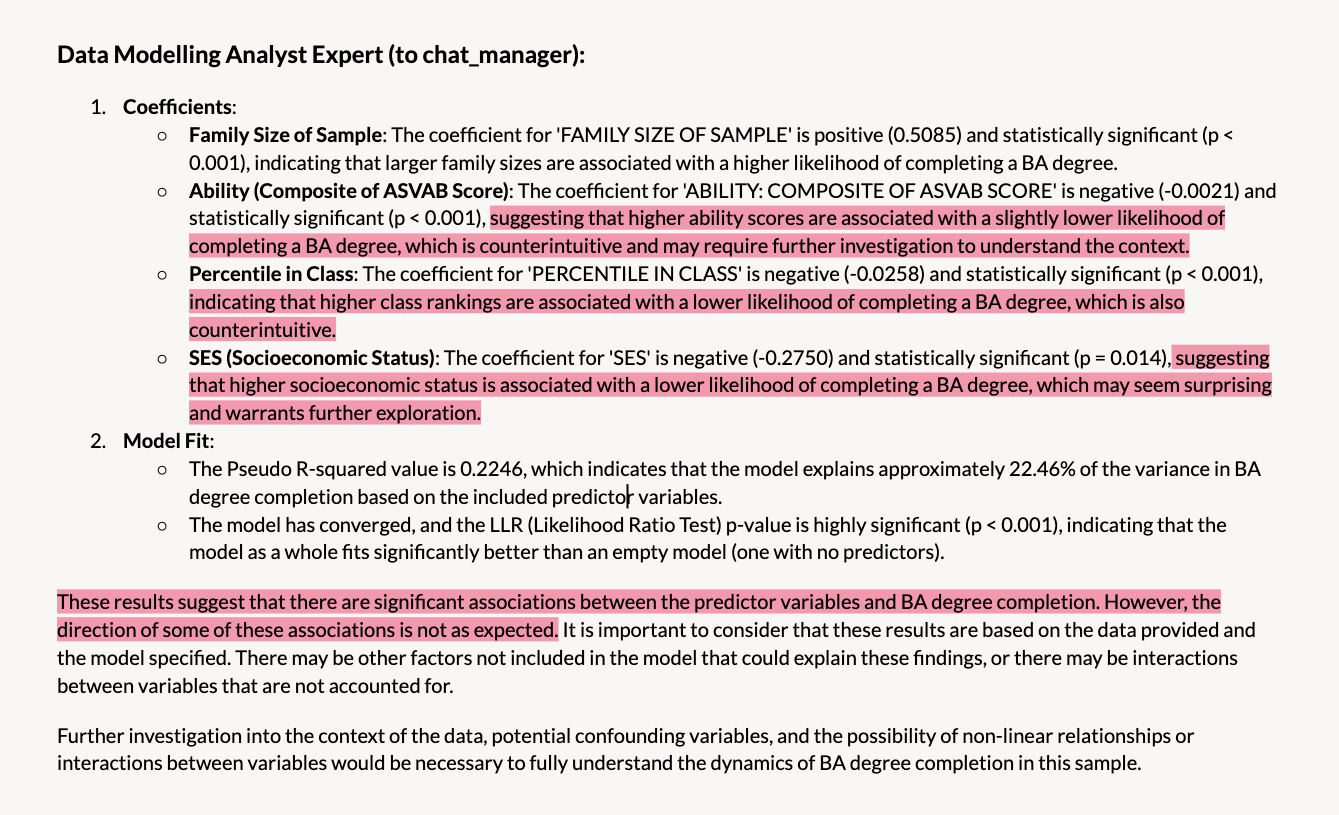}
    \caption{\textbf{Background:} We inverted the status of graduation in the 1979 follow-up wave of the National Longitudinal Survey dataset to verify the robustness of the \osod{}. We asked the usual query: how social background affects degree completion to \osod{}; \textbf{This figure:} The system correctly detects and communicates the surprising result (inverted trend) to the user (highlighted in red). An ideal data-discovery system should have the ability to detect and flag surprises in data.}
    \label{fig:enter-label}
\end{figure*}



\end{document}